\documentclass[10pt,journal,cspaper,compsoc,onecolumn]{IEEEtran}

\usepackage{setspace}
\usepackage{graphicx} 
\usepackage{subfigure} 

\usepackage{natbib}

\usepackage{algorithm}
\usepackage{algorithmic}
\usepackage[bottom]{footmisc}
\usepackage{algorithm}
\usepackage{algorithmic}
\usepackage{soul}
\usepackage{color}
\long\def\comment#1{}

\ifCLASSOPTIONcompsoc
\else
\fi
\ifCLASSINFOpdf
\else
\fi
\usepackage{url}

\begin{document}

%
\title{An Efficient Model Selection for Gaussian Mixture Model in a Bayesian Framework}
%
%
%
%

\author{Ji~Won~Yoon
\IEEEcompsocitemizethanks{\IEEEcompsocthanksitem J. Yoon is with the Center for Information Security 
Technology (CIST), Korea University, Korea.\protect\\
E-mail: $jiwon\_yoon@korea.ac.kr$
}
\thanks{}}

%
%

\markboth{IEEE Signal Processing Letters,~Vol.~xx, No.~xx, xxx ~2013}%
{Yoon \MakeLowercase{\textit{et al.}}: IEEE Signal Processing Letters}
%

 \maketitle

\begin{abstract} 
In order to cluster or partition data, we often use Expectation-and-Maximization (EM) or Variational approximation with a Gaussian Mixture Model (GMM), which  is a parametric probability density function represented as a weighted sum of $\hat{K}$ Gaussian component densities. However, model selection to find underlying $\hat{K}$ is one of the key concerns in GMM clustering, since we can obtain the desired clusters only when $\hat{K}$ is known. In this paper, we propose a new model selection algorithm to explore $\hat{K}$ in a Bayesian framework. The proposed algorithm builds the density of the model order which any information criterions such as AIC and BIC basically fail to reconstruct. In addition, this algorithm reconstructs the density quickly as compared to the time-consuming Monte Carlo simulation.
\end{abstract} 

\section{Introduction}

The Gaussian Mixture Model (GMM) is a well-known approach for clustering data. GMM is a parametric probability density function represented as a weighted sum of $K$ Gaussian component densities. Given an assumed model $\mathcal{M}_{K}$, the GMM takes the form 
\begin{equation}
p({\bf y}|\mathcal{M}_{K}) = \sum_{k=1}^{K}\pi_{k} p({\bf y}|\mu_{k}, {\bf Q}_{k})
\label{eq: Gaussian Mixture Model}
\end{equation} 
where ${\bf y}$ is a set of $N$ measurements (observations) and it has a multivariate ($d$-dimensional) continuous valued form. Here, $\pi_{k}$ and $p({\bf y}|\mu_{k}, {\bf Q}_{k})$ represent the mixture weight and the Gaussian density of the $k$-th component respectively. Each component density has the multivariate Gaussian function $p({\bf y}|\mu_{k}, {\bf Q}_{k}) = \mathcal{N}\left({\bf y}; \mu_{k}, {\bf Q}_{k}^{-1}\right)$ with mean $\mu_{k}$ and the covariance ${\bf Q}_{k}^{-1}$ of the $k$th component. Further, the sum of the non-negative weights is one, i.e $\sum_{k=1}^{K}\pi_{k}=1$ and $\pi_{k}\geq 0$. In this parameterized form, we can collectively represent hidden variables by ${\bf x}_{k}=( \pi_{k}, \mu_{k}, {\bf Q}_{k} )$ for $k=1, \cdots, K$. Now, our interest is to reconstruct the posterior distribution $p({\bf x}_{1:K}|{\bf y})$ given a model assumption that the Gaussian Mixture Model has $K$ components.

Let $K^{*}$ be the optimal number of Gaussian components, where $K^{*}=\arg_{K}\max p(K|{\bf y})$. It is known that if $K^{*}$ is known, the desired ${\bf x}_{1:K^{*}}$ is straightforwardly estimated by Variational approximation or classic Expectation-and-Maximization algorithm (EM). However, in general the optimal number of components $K^{*}$ is not known and therefore it is rather difficult to estimate hidden parameters ${\bf x}_{1:K^{*}}$. This is because the dimension of ${\bf x}_{1:K}$ is changing with varying $K$. Generically, the model selection problem for GMM involves finding the optimal $K^{*}$ by $K^{*} = \arg_{K} \max p(K|{\bf y})$. There are many studies in the literature that have addressed the model order estimation for GMM \cite{Fruhwirth-Schnatter_2007,Kadane_Lazar_2004,BISHEM01,Roberts98:ModelSelectionGMM,Constantinopoulos_bayesianfeature}. For instance, Keribin et al. \cite{Keribin_2000} estimated the number of components for mixture models using a maximum penalized likelihood. \emph{Information criterion}s have also been applied to GMM model selection, such as  AIC \cite{Aitkin85:ModelSelection4MixtureModel}, BIC \cite{Fraley98:numClust} and the Entropy criterion \cite{Celeux_Soromenho_1996}. In the Monte Carlo simulation, Richardson et al. developed the inference of GMM model selection using the reversible jump Markov chain Monte Carlo (RJMCMC) \cite{Richardson97:GMM}. Recently, Nobile et al. introduced an efficient clustering algorithm, the so called \emph{allocation sampler}  \cite{Nobile_Fearnside_2010}, which basically infers the model order and clusters using Monte Carlo in an efficient marginalized proposal distribution.

\section{Statistical Background}
\label{section: Statistical Background}

\subsection{Integrated Nested Laplace Approximation (INLA)}
Suppose that we have a set of hidden variables ${\bf f}$ and a set of observations ${\cal Y}$. Integrated Nested Laplace Approximation (INLA) \cite{Rue09:INLA} approximates the marginal posterior $p({\bf f|{\cal Y}})$ by
\begin{eqnarray}
p({\bf f}|{\cal Y}) &=& \int p({\bf f}|{\cal Y}, \theta)p(\theta|{\cal Y})d\theta\cr
&\approx& \int \tilde{p}({\bf f}|{\cal Y}, \theta)\tilde{p}(\theta|{\cal Y})d\theta\cr
&\approx& \sum_{\theta_{i}}\tilde{p}({\bf f}|{\cal Y}, \theta)\tilde{p}(\theta|{\cal Y})\Delta_{\theta_{i}}\nonumber \textrm{ where}
\end{eqnarray}
\begin{equation}
\tilde{p}(\theta|{\cal Y}) \propto \left.\frac{p({\bf f}, {\cal Y}, \theta)}{p_{F}({\bf f}|{\cal Y}, \theta)}\right|_{{\bf f}={\bf f}^{*}(\theta)}= \left.\frac{p({\cal Y}|{\bf f}, \theta)p({\bf f}|\theta)p(\theta)}{p_{F}({\bf f}|{\cal Y}, \theta)}\right|_{{\bf f}={\bf f}^{*}(\theta)}.
\label{eq: marginal posterior of theta in INLA}
\end{equation}
Here, $F$ denotes a simple functional approximation close to $p({\bf f}|{\cal Y}, \theta)$, as in Gaussian approximation, and ${\bf f}^{*}(\theta)$ is a value of the functional approximation. For the simple Gaussian approximation case, the proper choice of ${\bf f}^{*}(\theta)$ is the mode of Gaussian approximation of $p_{G}({\bf f}|{\cal Y}, \theta)$. Given the log of posterior, we can calculate a mode ${\theta}^{*}$ and its Hessian matrix ${\bf H}_{\theta}^{*}$ via quasi-Newton style optimization: $\theta^{*} = \arg_{\theta}\max \log \tilde{p}(\theta|{\cal Y})$, and for ${\bf H}^{*}_{\theta}$, we do a grid search from the mode in all directions until the log $\tilde{p}({\theta}^{*}|{\cal Y})-\log\tilde{p}(\theta|{\cal Y})>\varphi$ for a given threshold $\varphi$.

\section{Proposed Approach}
\label{section: Proposed Approach}
In this study, we extend our previous work, which addressed model selection for the K-nearest neighbour classifier using the {\bf K-OR}der {\bf E}stimation {\bf A}lgorithm (KOREA) \cite{Yoon13:PKNN_arXiv}, to resolve the model selection problem in clustering domains using KOREA. Our proposed algorithm reconstructs the distribution of the number of components using Eq. (\ref{eq: marginal posterior of theta in INLA}). 

\subsection{Obtaining the optimal number of components}
Let ${\bf y}$ denote a set of observations and let ${\bf x}_{1:K}$ be a set of the model parameters given a model order $K$. The first step of our algorithm is to estimate the optimal number of components, $K^{*}$: $K^{*} = \arg_{K}\max p(K|{\bf y})$. According to Eq. (\ref{eq: marginal posterior of theta in INLA}), we can obtain an approximated marginal posterior distribution by
\begin{equation}
\tilde{p}(K|{\bf y}) \propto \left.\frac{p({\bf y}, {\bf x}_{1:K}, K)}{p_{F}({\bf x}_{1:K}|{\bf y}, K)}\right|_{{\bf x}_{1:K}(K)={\bf x}_{1:K}^{*}(K)}.
\label{eq: the general format of KOREA}
\end{equation}
This equation has the property that $K$ is an integer variable, while $\theta$ of Eq. (\ref{eq: marginal posterior of theta in INLA}) is, in general, continuous variables. By ignoring this difference, we can still use a quasi-Newton method to obtain optimal $K^{*}$ efficiently. Alternatively, we can also calculate some potential candidates between $1$ and $K_{\max}$ if $K_{\max}$ is not too large. Otherwise, we may still use the \emph{quasi-Newton} style algorithm with a rounding operator that transforms a real value to an integer for $K$.

\subsection{Bayesian Model Selection for GMM}
In the GMM model of Eq. (\ref{eq: Gaussian Mixture Model}), we have four different types of hidden variables for the profile of the components: mean ($\mu_{1:K}$), precision (${\bf Q}_{1:K}$), the weights ($\pi_{1:K}$) of the component and an unknown number of components $K$. Therefore, given  Eq. (\ref{eq: the general format of KOREA}), we can make the mathematical form: $\tilde{p}(K|{\bf y})\propto \left.\frac{p({\bf y}|{\bf x}_{1:K})p({\bf x}_{1:K})p(K)}{p_{F}({\bf x}_{1:K}|{\bf y}, K)}\right|_{{\bf x}_{1:K}={\bf x}_{1:K}^{*}(K)}$, where ${\bf y}={\bf y}_{1:N}$ and ${\bf x}_{1:K}=(\pi_{1:K}, \mu_{1:K}, {\bf Q}_{1:K})$. However, it is rather difficult to obtain the approximated distribution $p_{F}({\bf x}_{1:K}|{\bf y}, K)$ close to target distribution since there is no close form. Worse, it is infeasible to build a Hessian matrix via a quasi-Newton method since it is extremely slow when the dimension of ${\bf x}$ is large and ${\bf Q}$ is not a vector but a matrix. Therefore, we introduce labeling indicator ${\bf z}$ to decompose the mixture model and apply the Variational approach \cite{Bishop06:MachineLearning} to obtain $p_{F}({\bf z}, {\bf x}_{1:K}|{\bf y}, K)$ such that $q({\bf z}, {\bf x}_{1:K})=p_{F}({\bf z}, {\bf x}_{1:K}|{\bf y}, K)$. Therefore, we re-define the problems by adding component indicators of observations ${\bf z}$. We finally obtain in a form similar to that of Eq. (\ref{eq: the general format of KOREA})
\begin{eqnarray}
\tilde{p}(K|{\bf y})&\propto& \left.\frac{p({\bf y}, {\bf z}, {\bf x}_{1:K}, K)}{p_{F}({\bf z}, {\bf x}_{1:K}|{\bf y}, K)}\right|_{({\bf z}, {\bf x}_{1:K})=({\bf z}, {\bf x}_{1:K})^{*}(K)}\cr
&=& \frac{p({\bf y}|{\bf z}^{*}, {\bf x}_{1:K}^{*})p({\bf z}^{*}|{\bf x}_{1:K}^{*})p({\bf x}_{1:K}^{*})p(K)}{q^{*}({\bf z}^{*}, {\bf x}_{1:K}^{*})}
\end{eqnarray}
where ${\bf z}^{*}$ and ${\bf x}_{1:K}^{*}$ are the mode of $q^{*}({\bf z}, {\bf x}_{1:K})$, which is an approximated posterior obtained by variational approximation. Here, $p(K)=\frac{\exp(-K)}{\sum_{j=1}^{K_{\max}}\exp(-j)}$.

\section{Evaluation}
\label{section: Evaluation}

In order to evaluate the performance of our proposed approach, we simulated our clustering algorithm using an artificial dataset and several real experimental datasets.

\begin{figure*}[t]
\centering
\begin{tabular}{cccc}
& (a) AIC & (b) BIC & (c) Our approach\cr
 $\hat{K}=1$ &
\includegraphics[height=1in, width=1.8in]{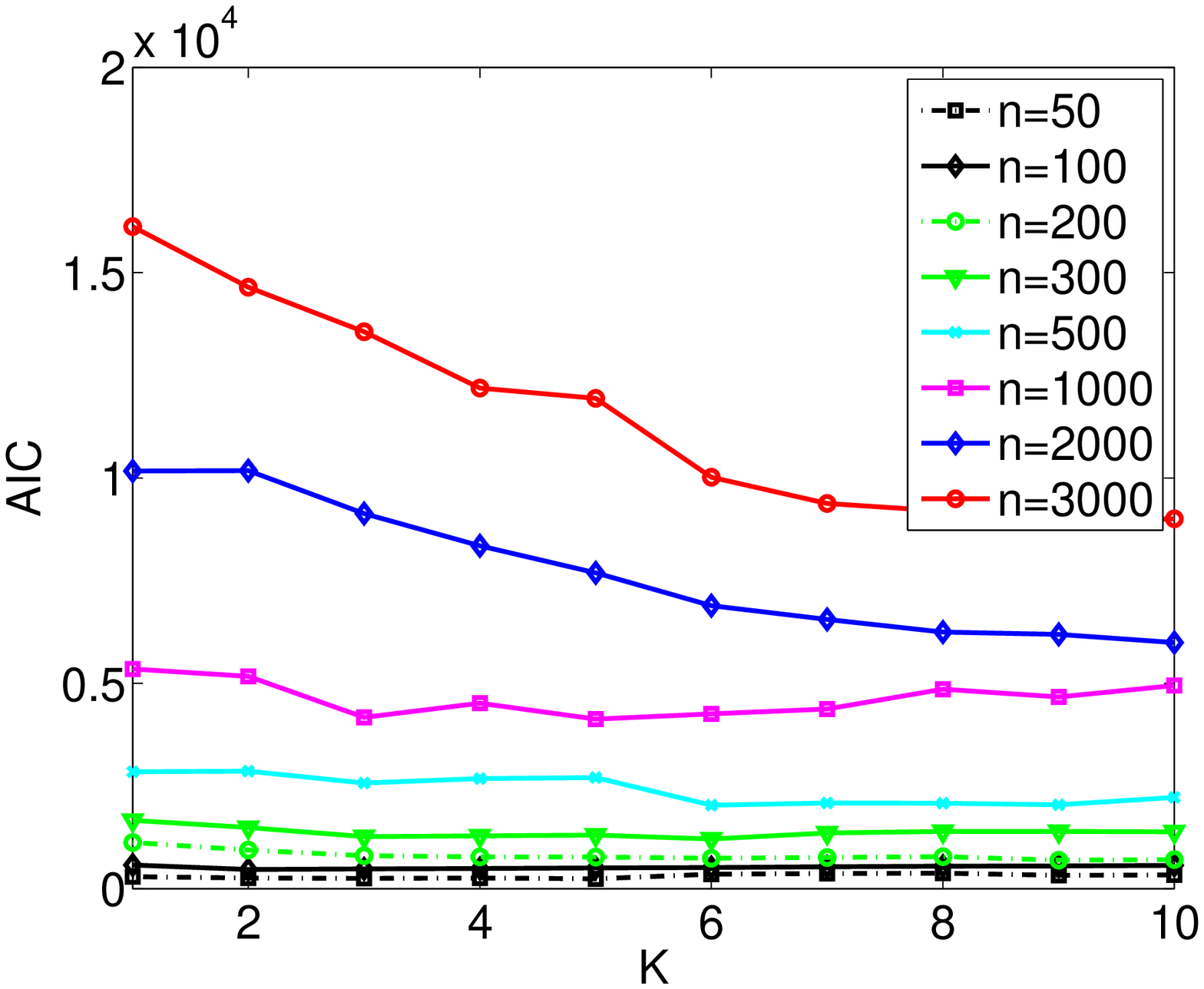}& 
\includegraphics[height=1in, width=1.8in]{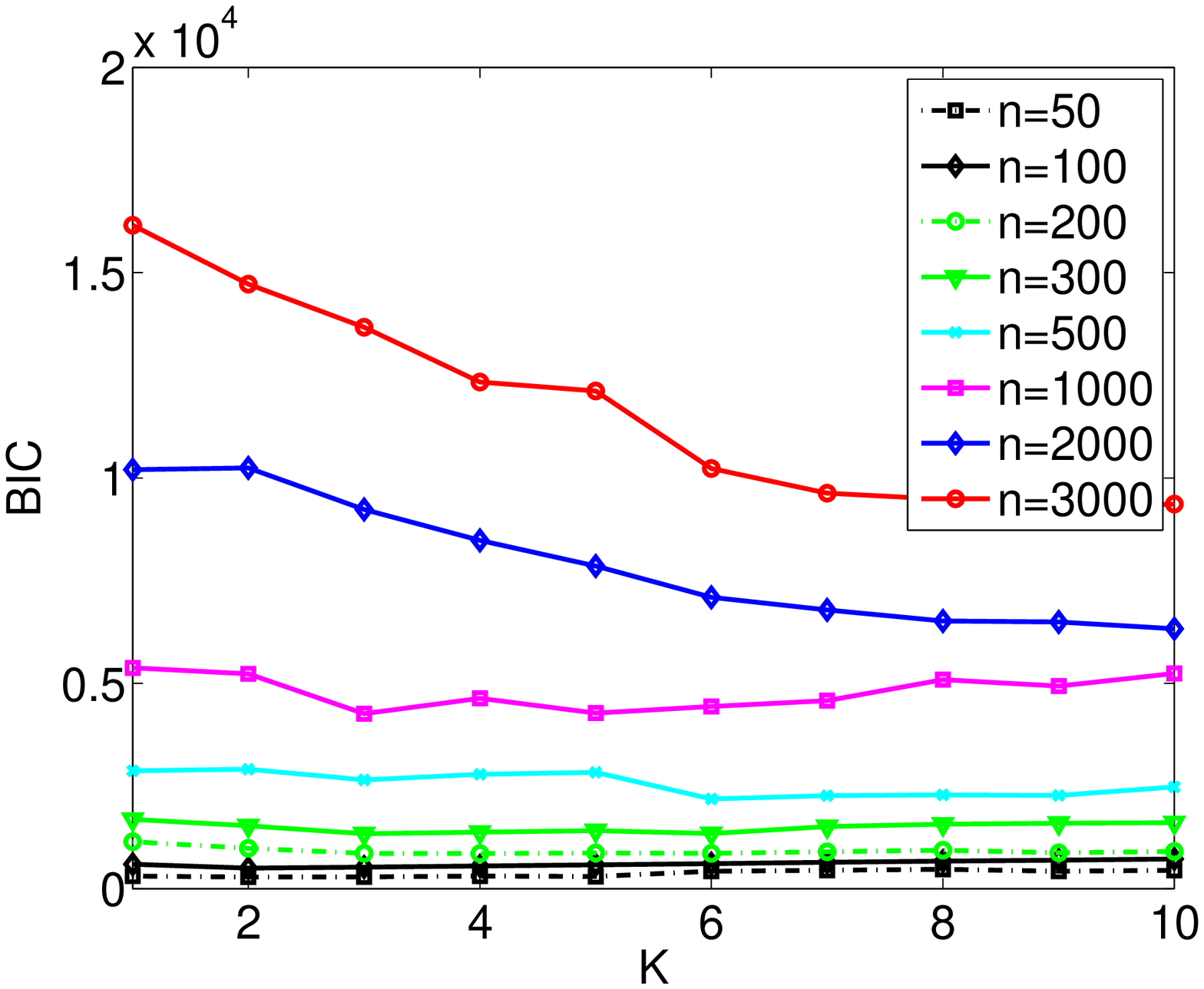}& 
\includegraphics[height=1in, width=1.8in]{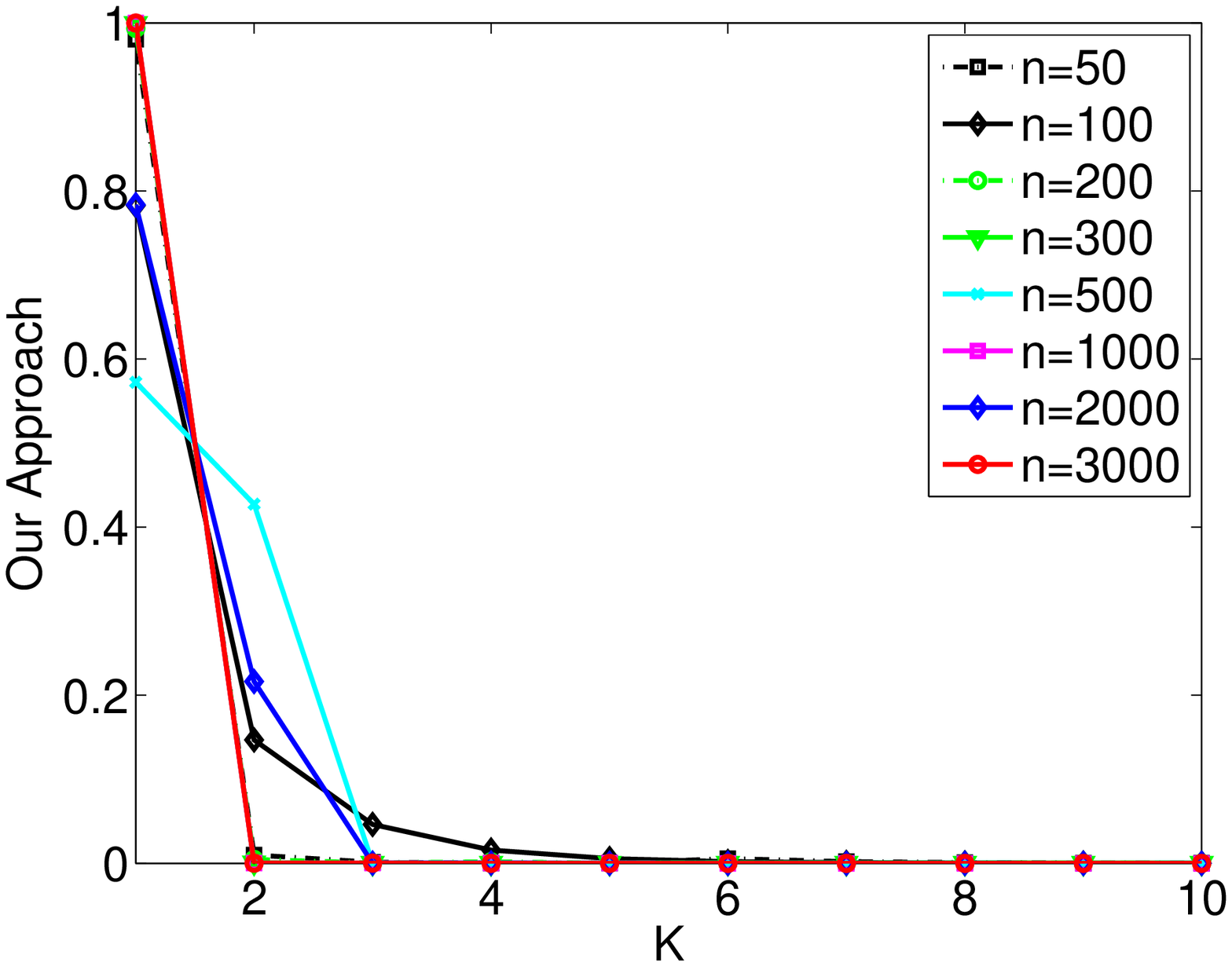}\cr
 $\hat{K}=2$ &
\includegraphics[height=1in, width=1.8in]{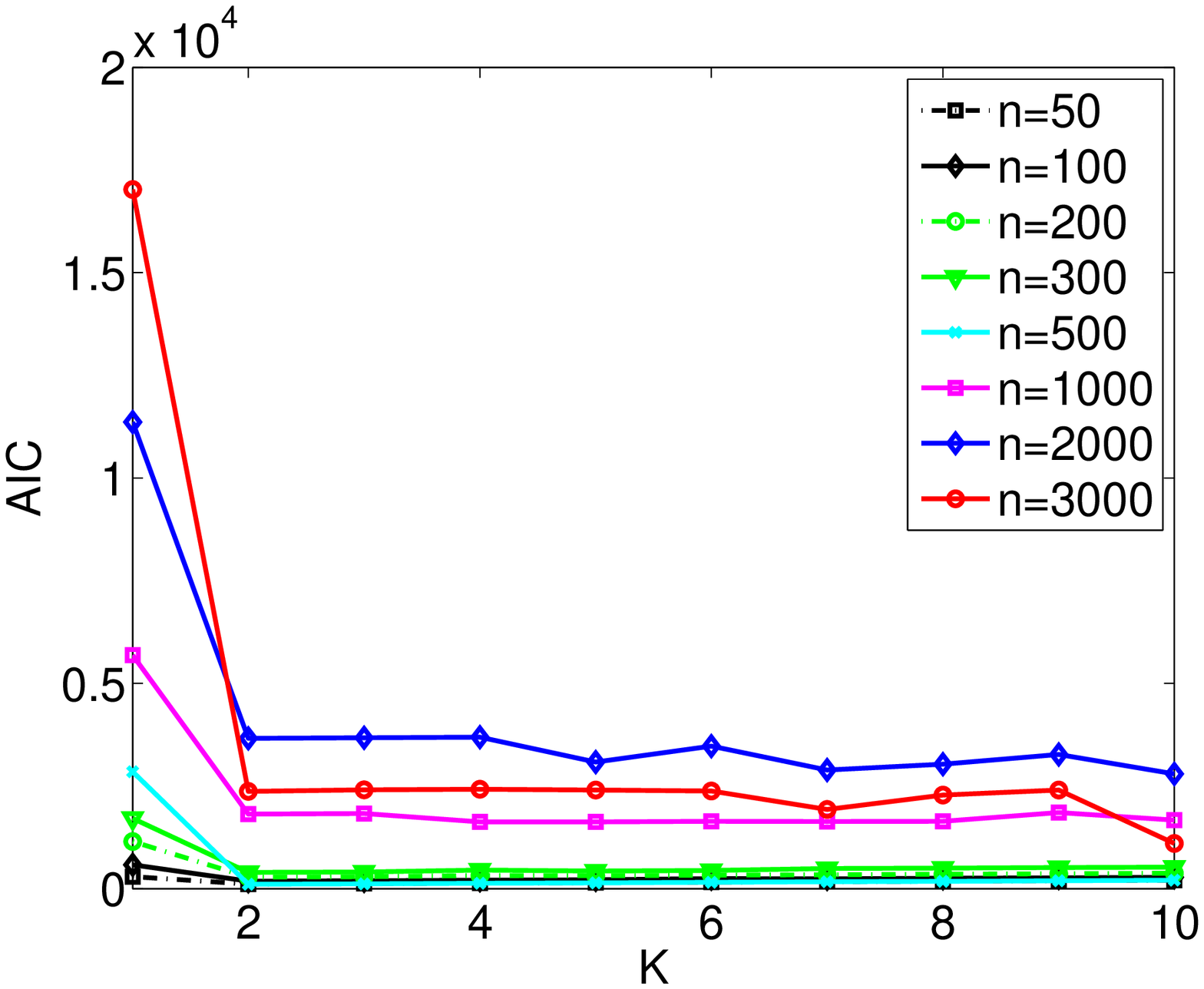}& 
\includegraphics[height=1in, width=1.8in]{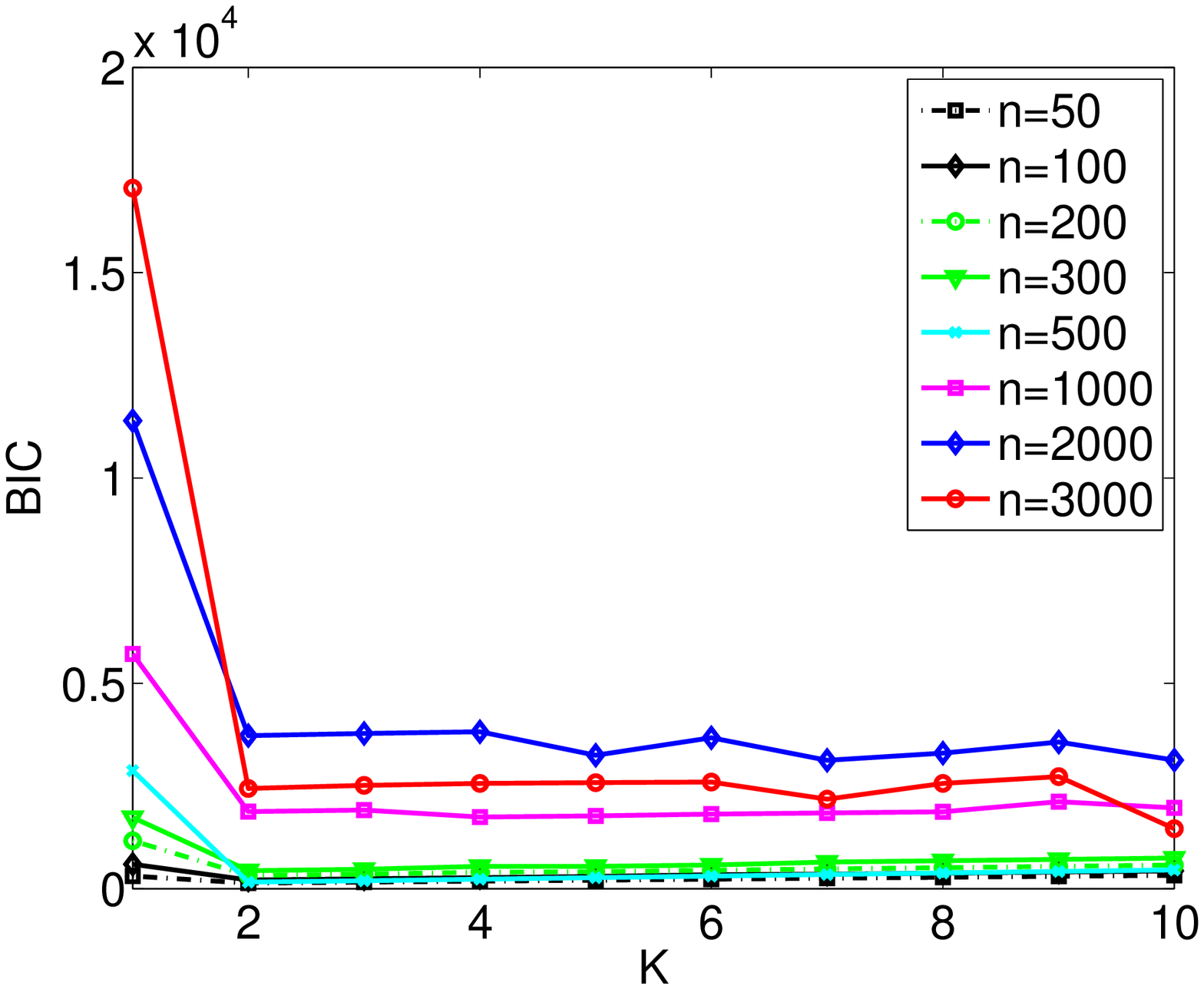}& 
\includegraphics[height=1in, width=1.8in]{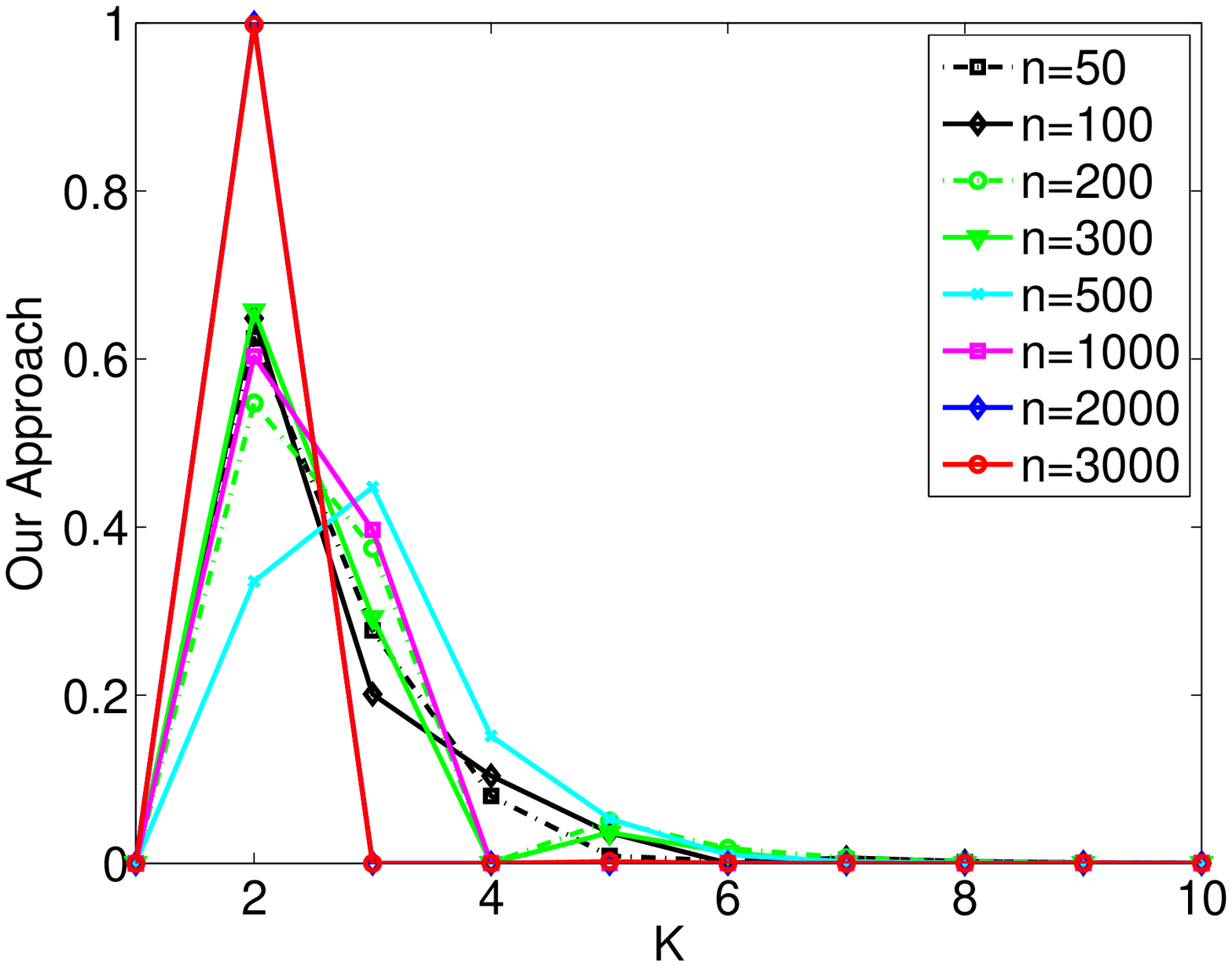}\cr
 $\hat{K}=3$ &
\includegraphics[height=1in, width=1.8in]{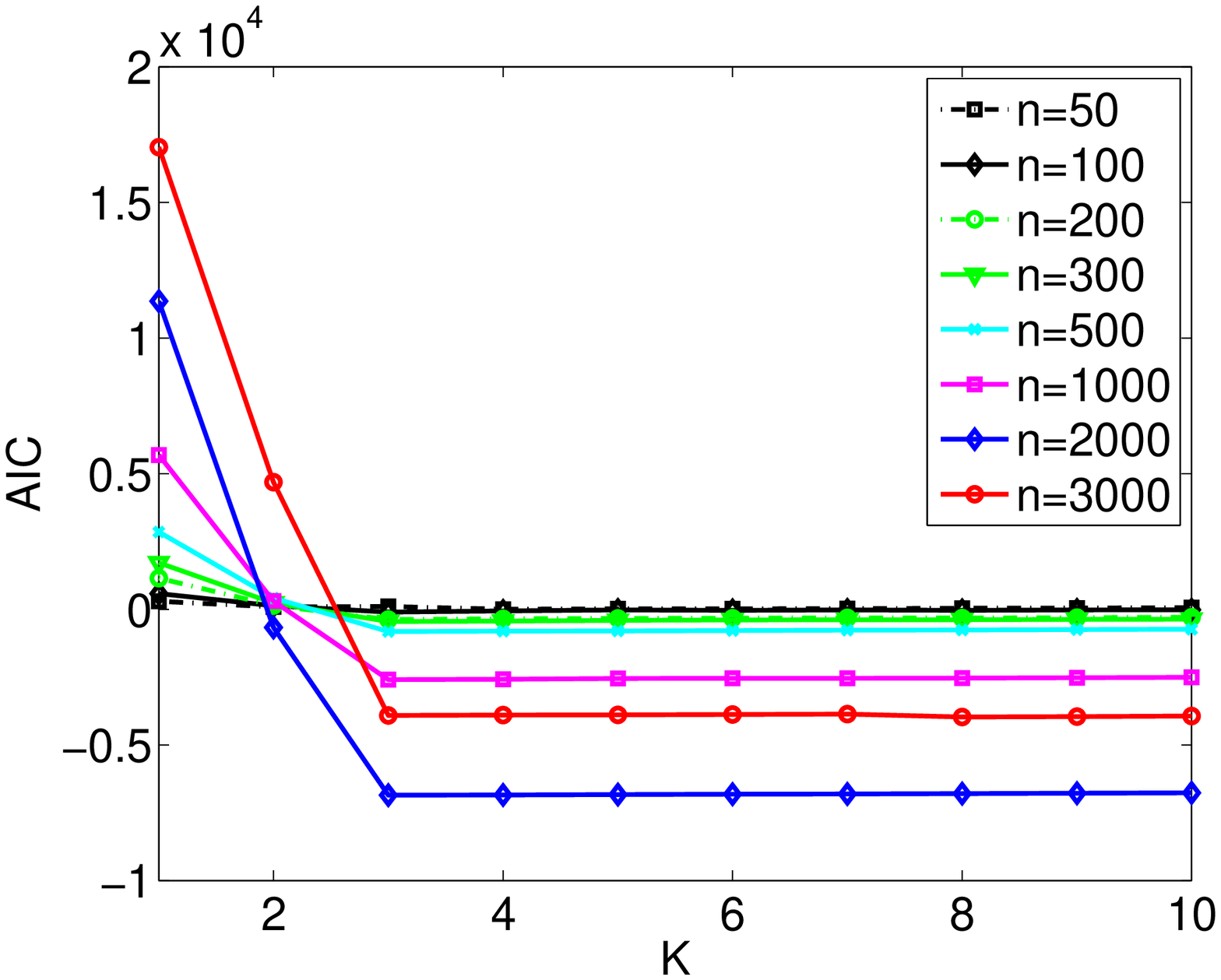}& 
\includegraphics[height=1in, width=1.8in]{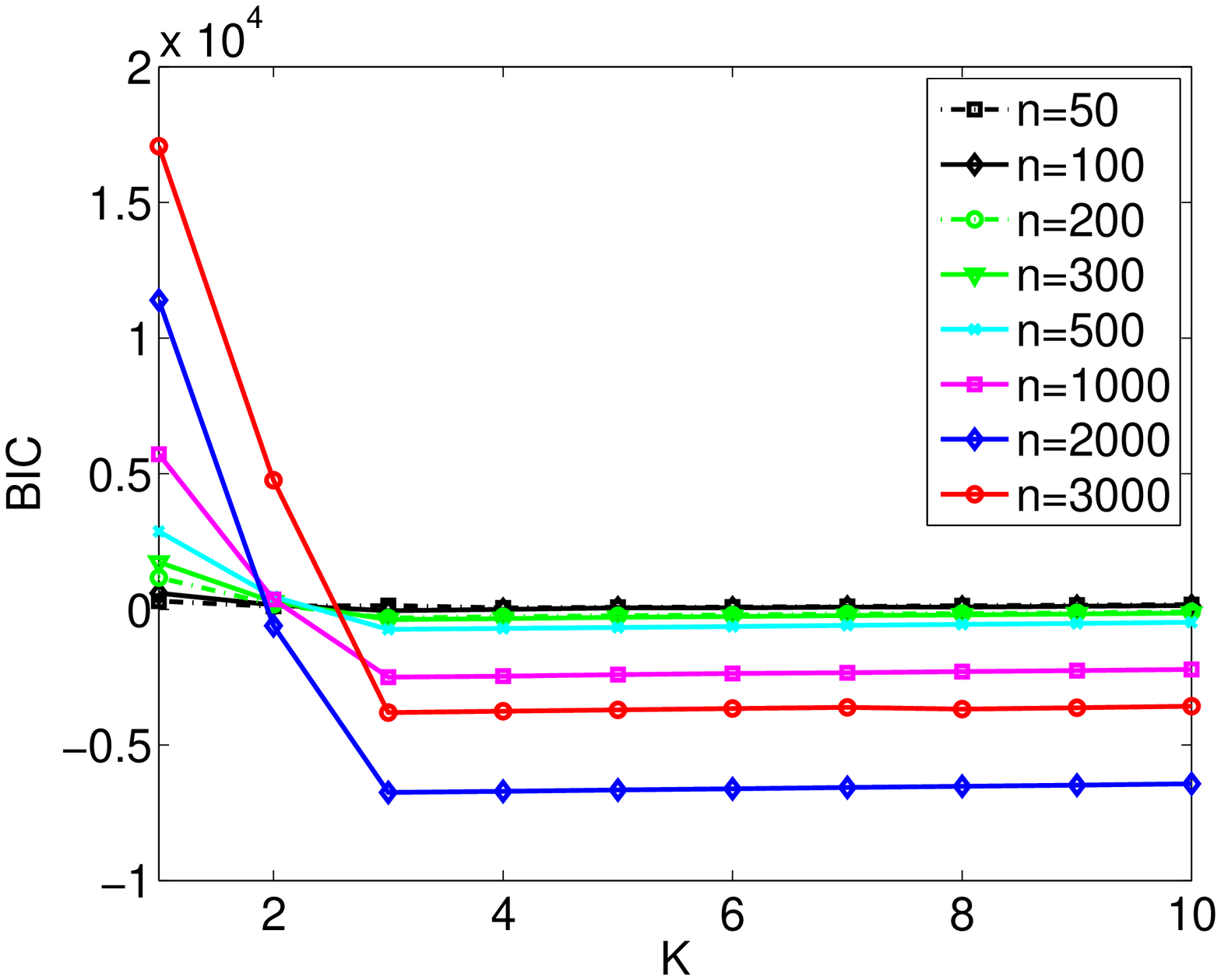}& 
\includegraphics[height=1in, width=1.8in]{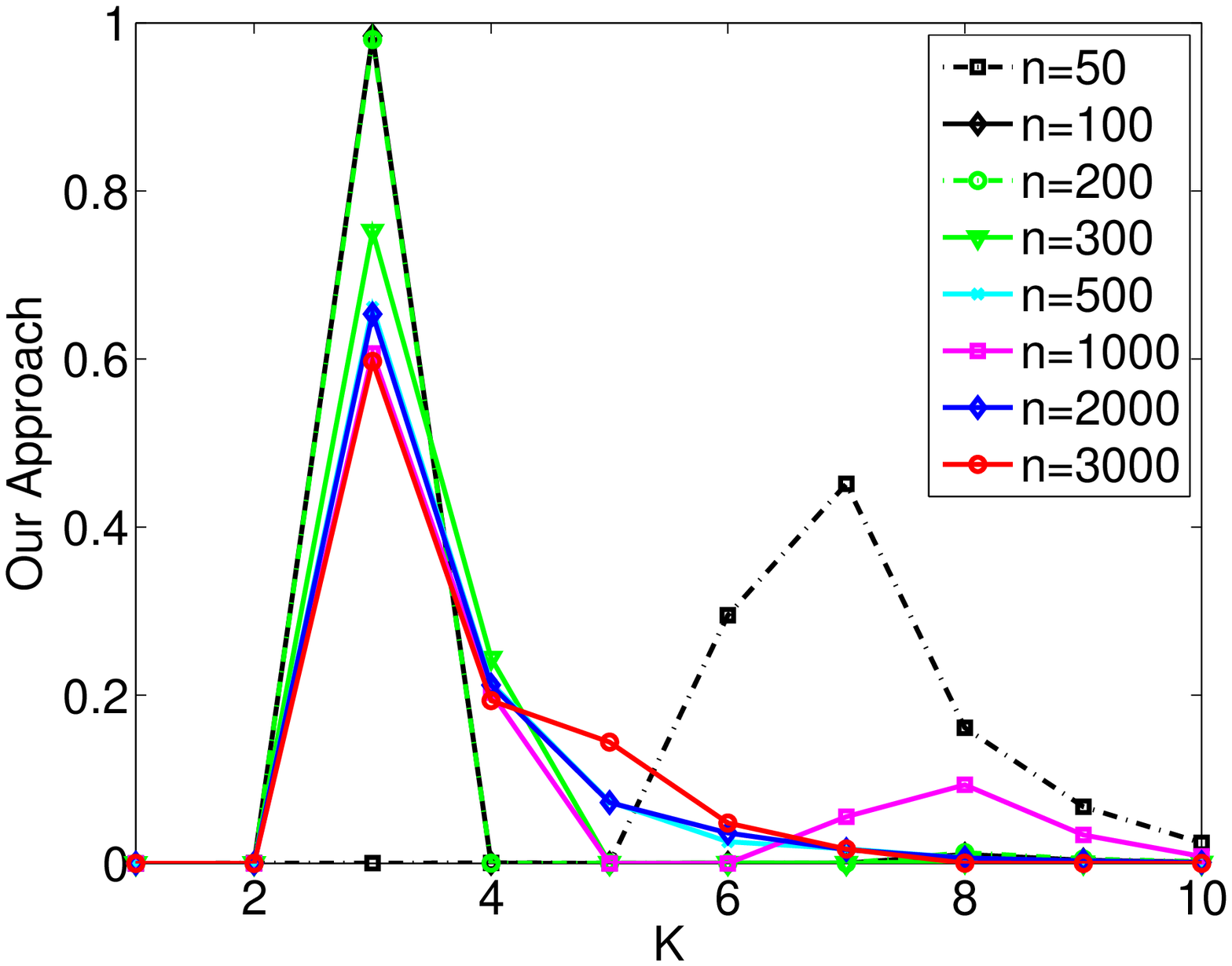}\cr
 $\hat{K}=4$ &
\includegraphics[height=1in, width=1.8in]{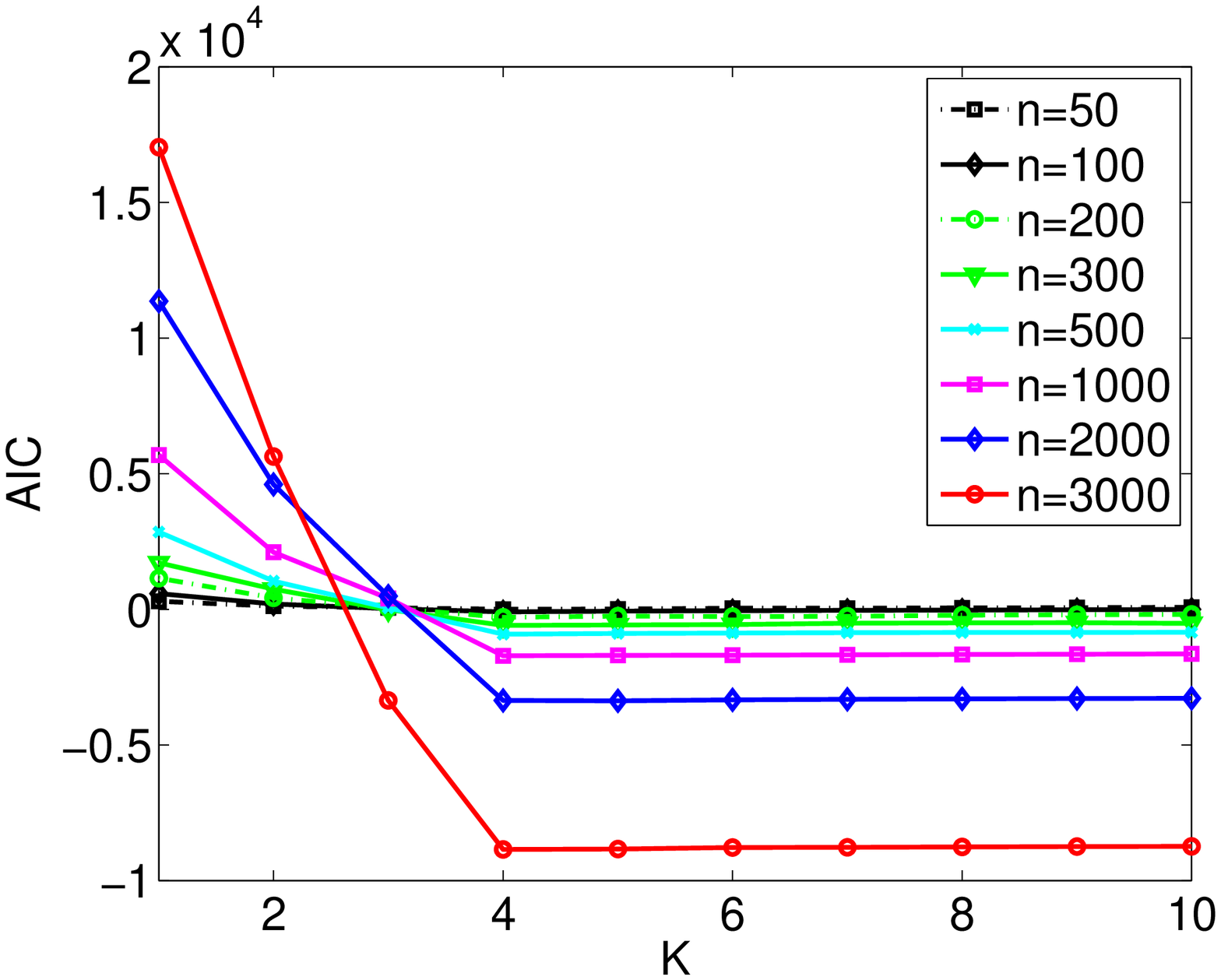}& 
\includegraphics[height=1in, width=1.8in]{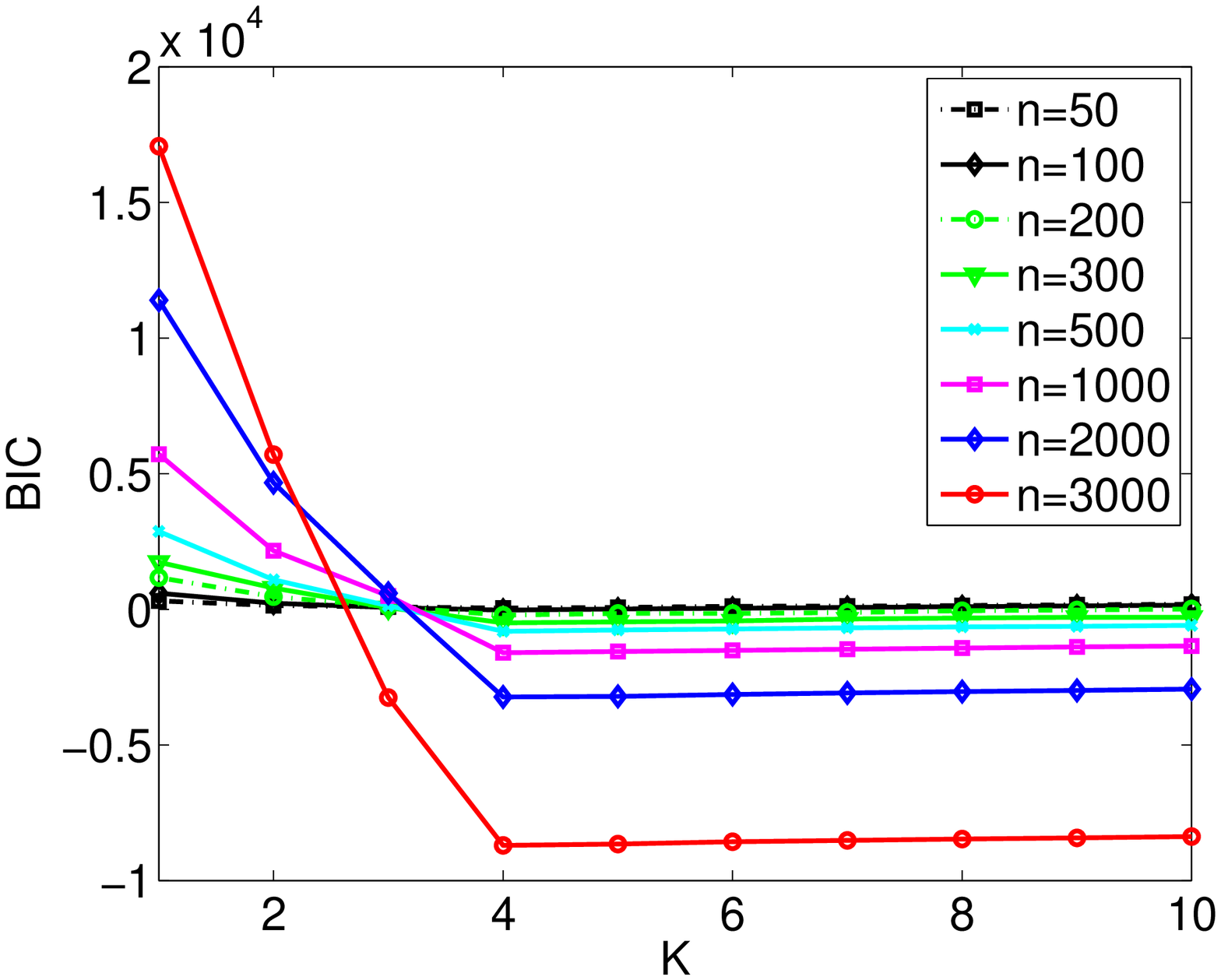}& 
\includegraphics[height=1in, width=1.8in]{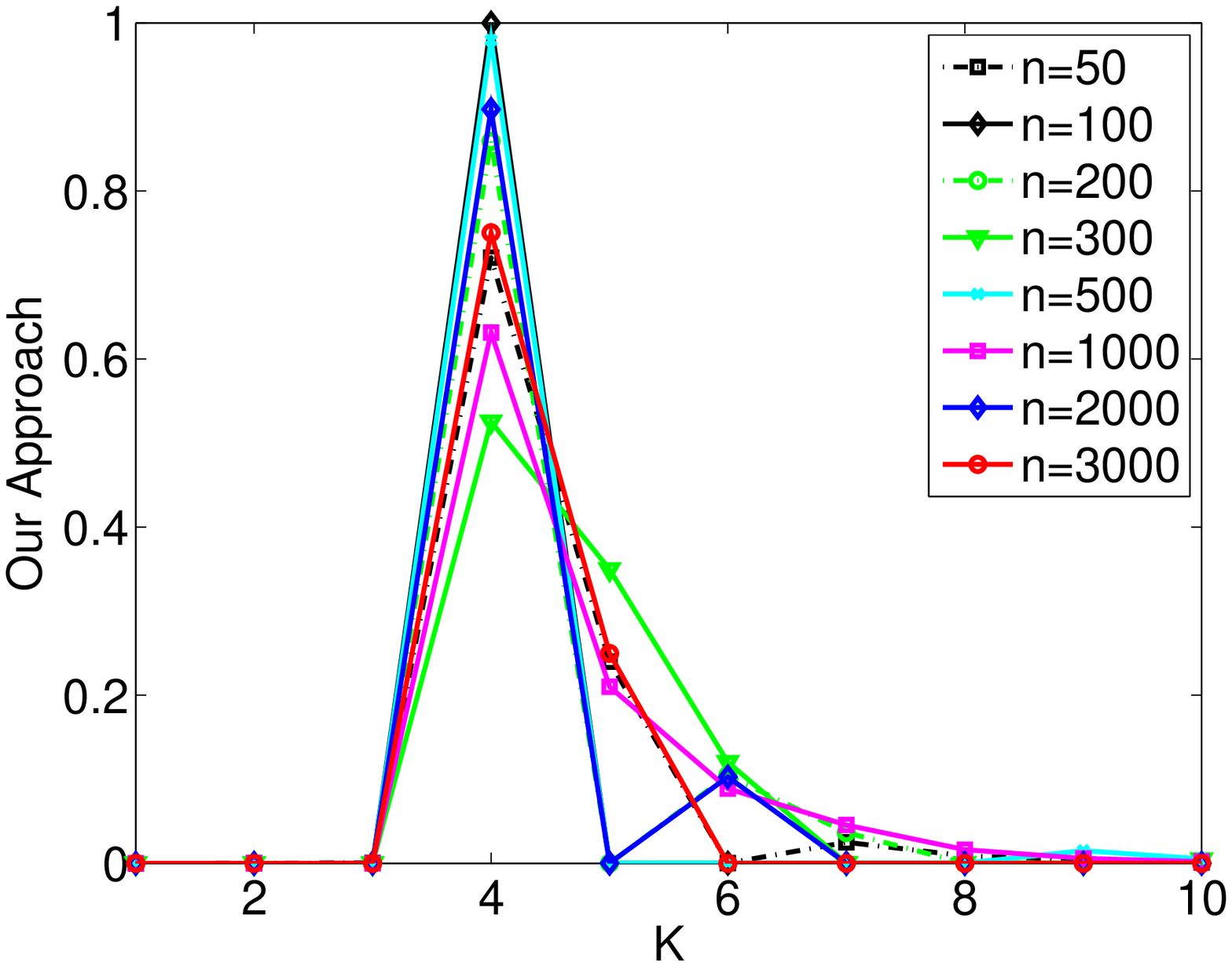}\cr
 $\hat{K}=5$ &
\includegraphics[height=1in, width=1.8in]{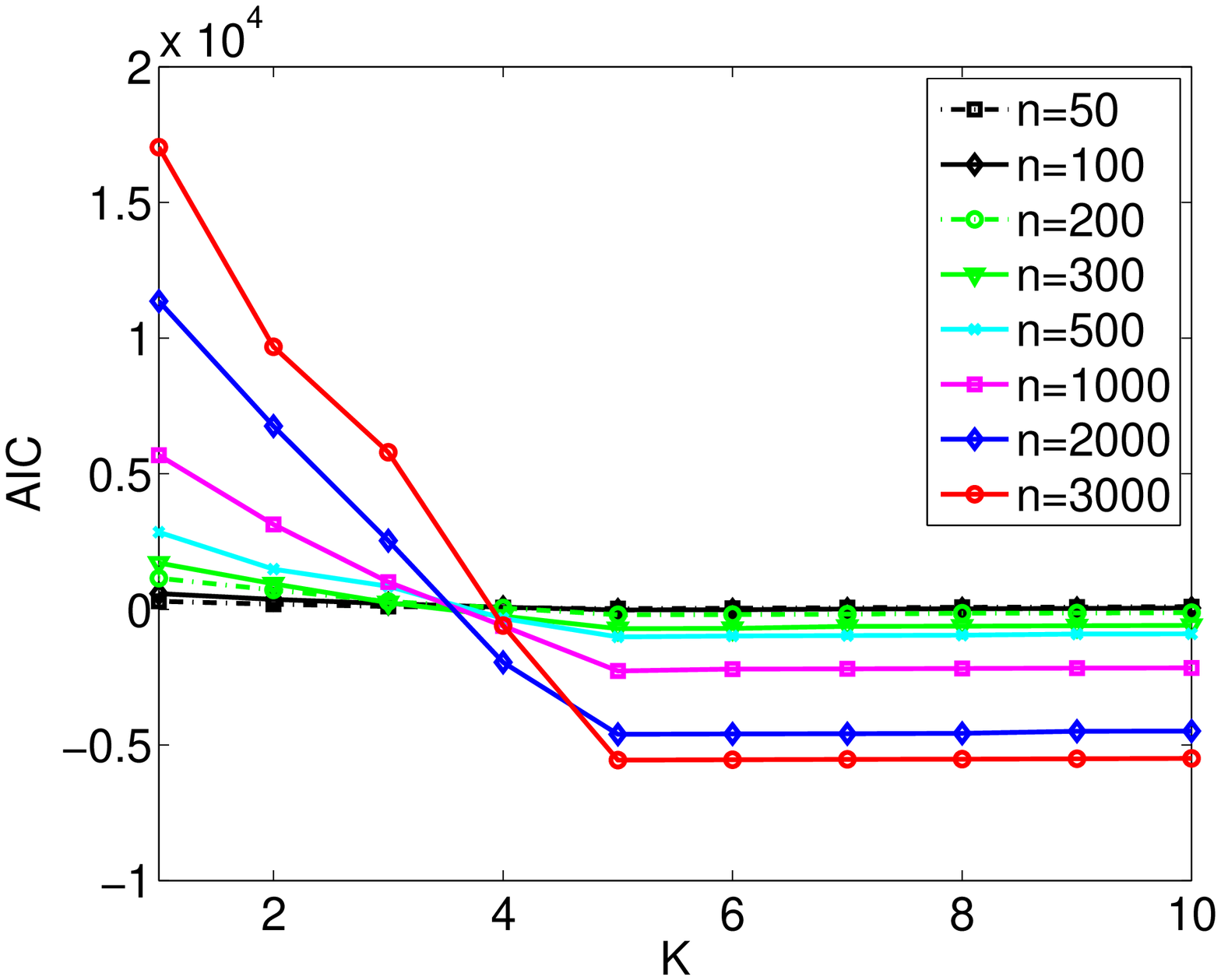}& 
\includegraphics[height=1in, width=1.8in]{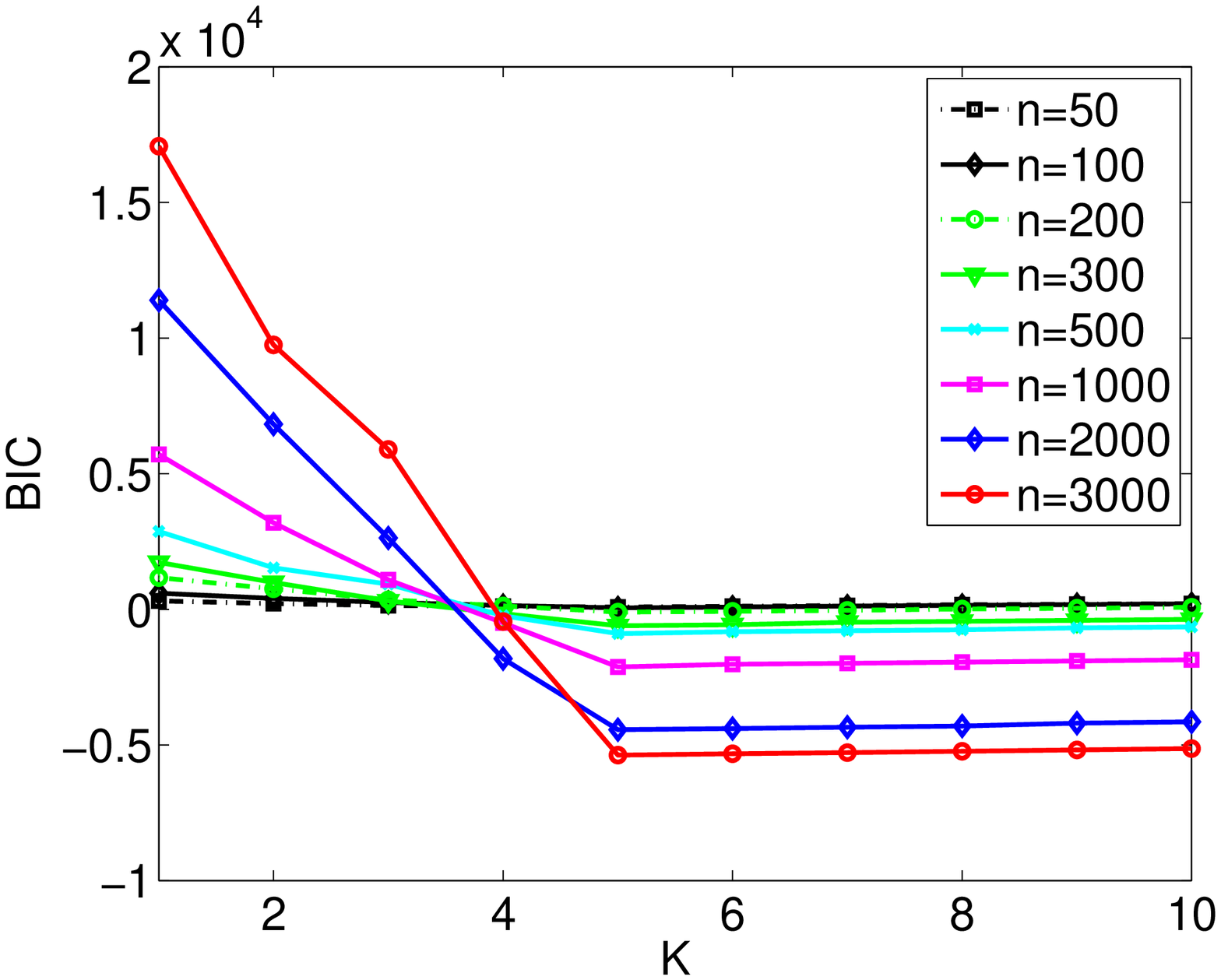}& 
\includegraphics[height=1in, width=1.8in]{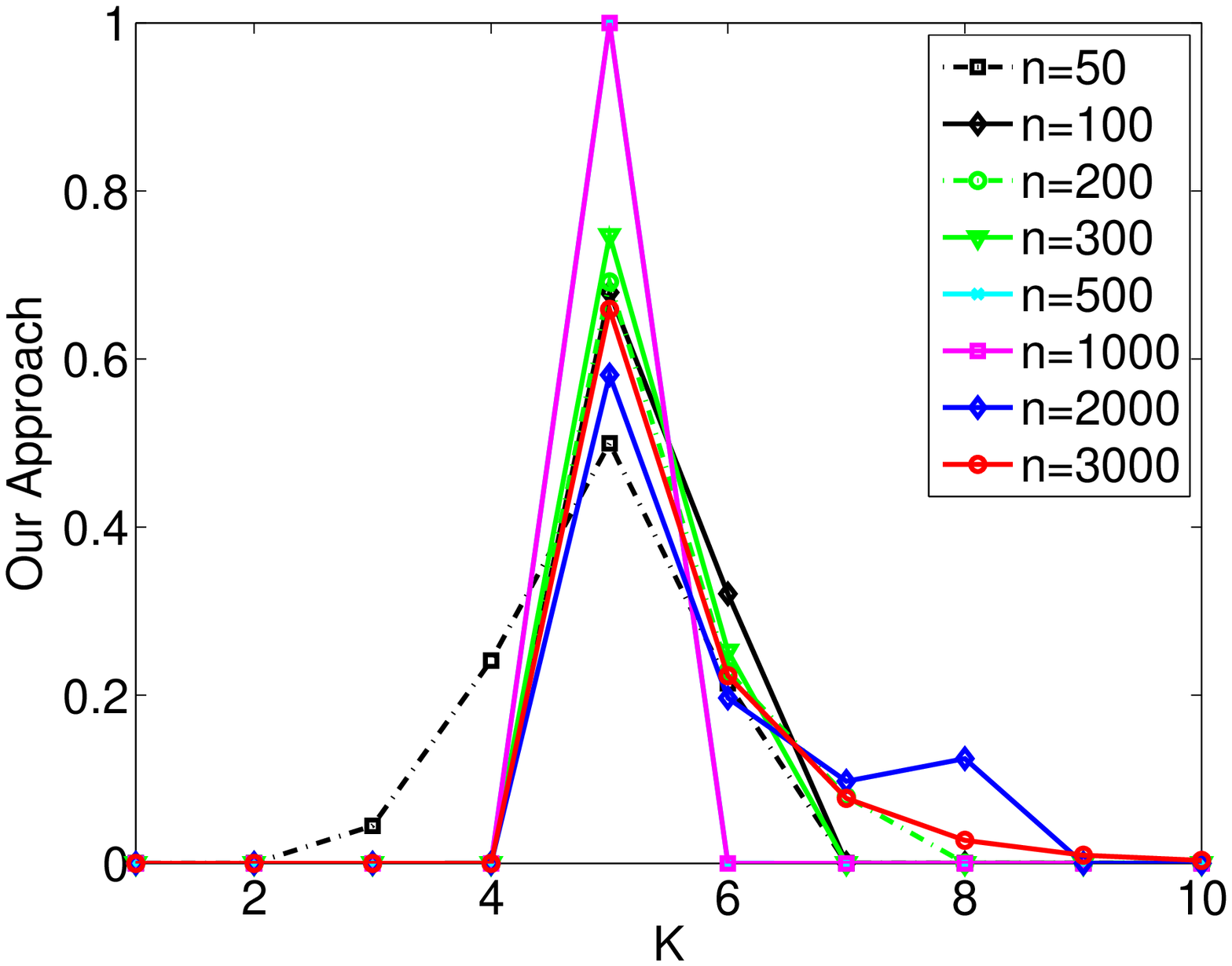}\cr
\end{tabular}
\caption{Clustered synthetic dataset where the number of clusters $K$ and the number of observations $N$ are varied: (a) AIC, (b) BIC, and (c) our proposed approach}
\label{fig: Clustered synthetic dataset with varying K}
\end{figure*}

\begin{figure*}[t]
\centering
\begin{tabular}{cccccc}
& $\hat{K}=1$ &  $\hat{K}=2$ & $\hat{K}=3$ & $\hat{K}=4$ & $\hat{K}=5$\cr
(a) Mean&
\includegraphics[scale=0.145]{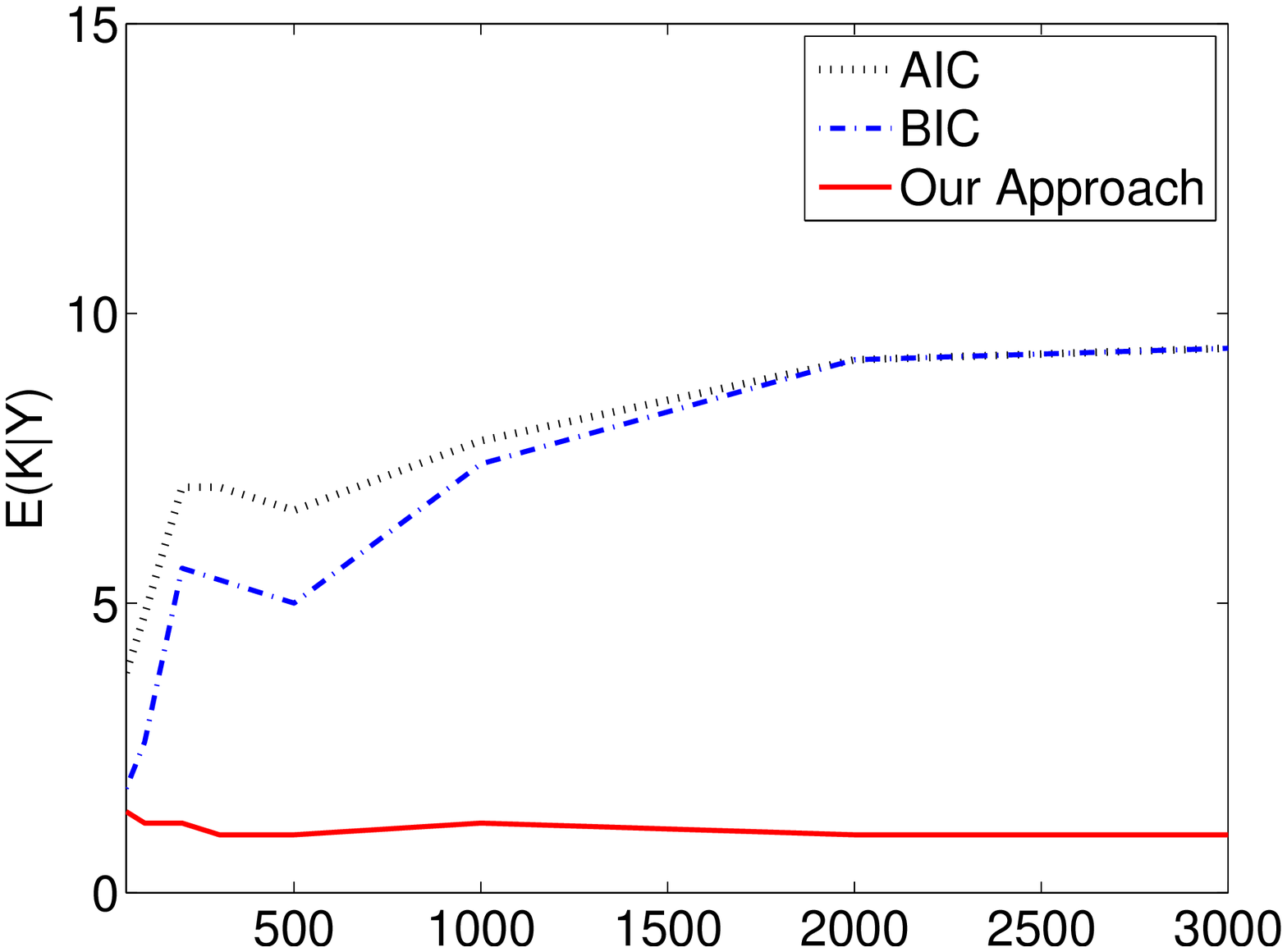}& 
\includegraphics[scale=0.145]{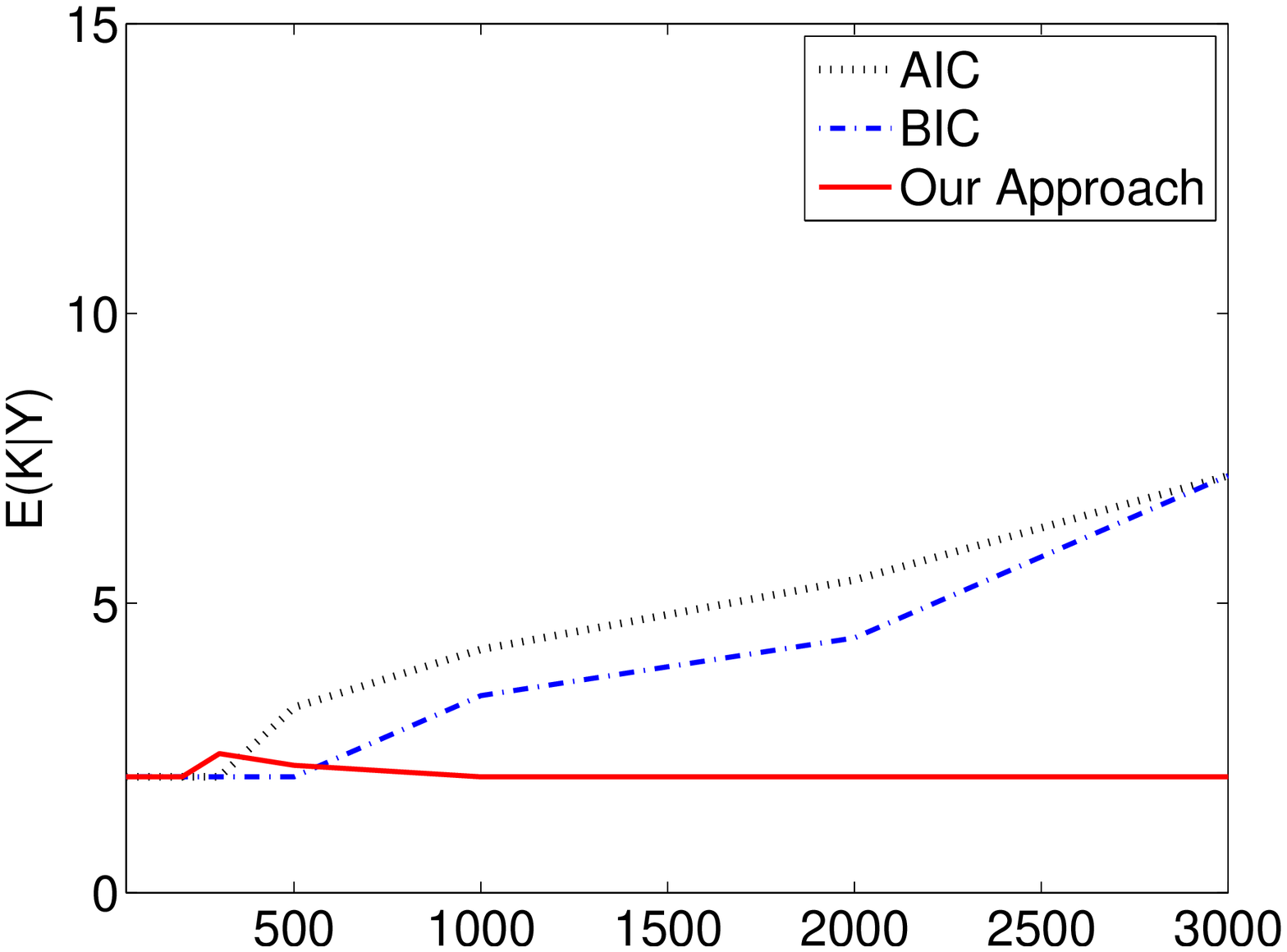}& 
\includegraphics[scale=0.145]{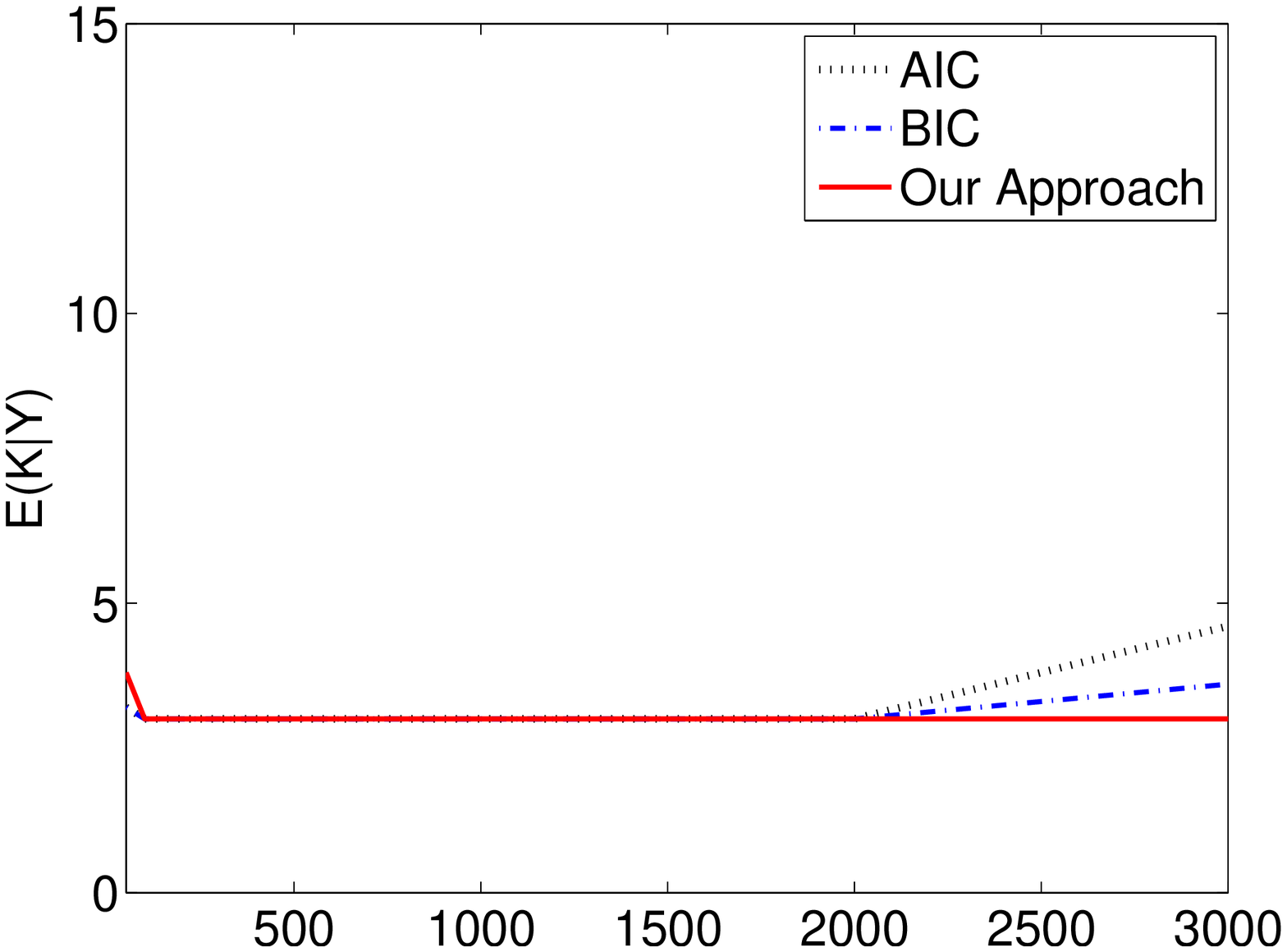}& 
\includegraphics[scale=0.145]{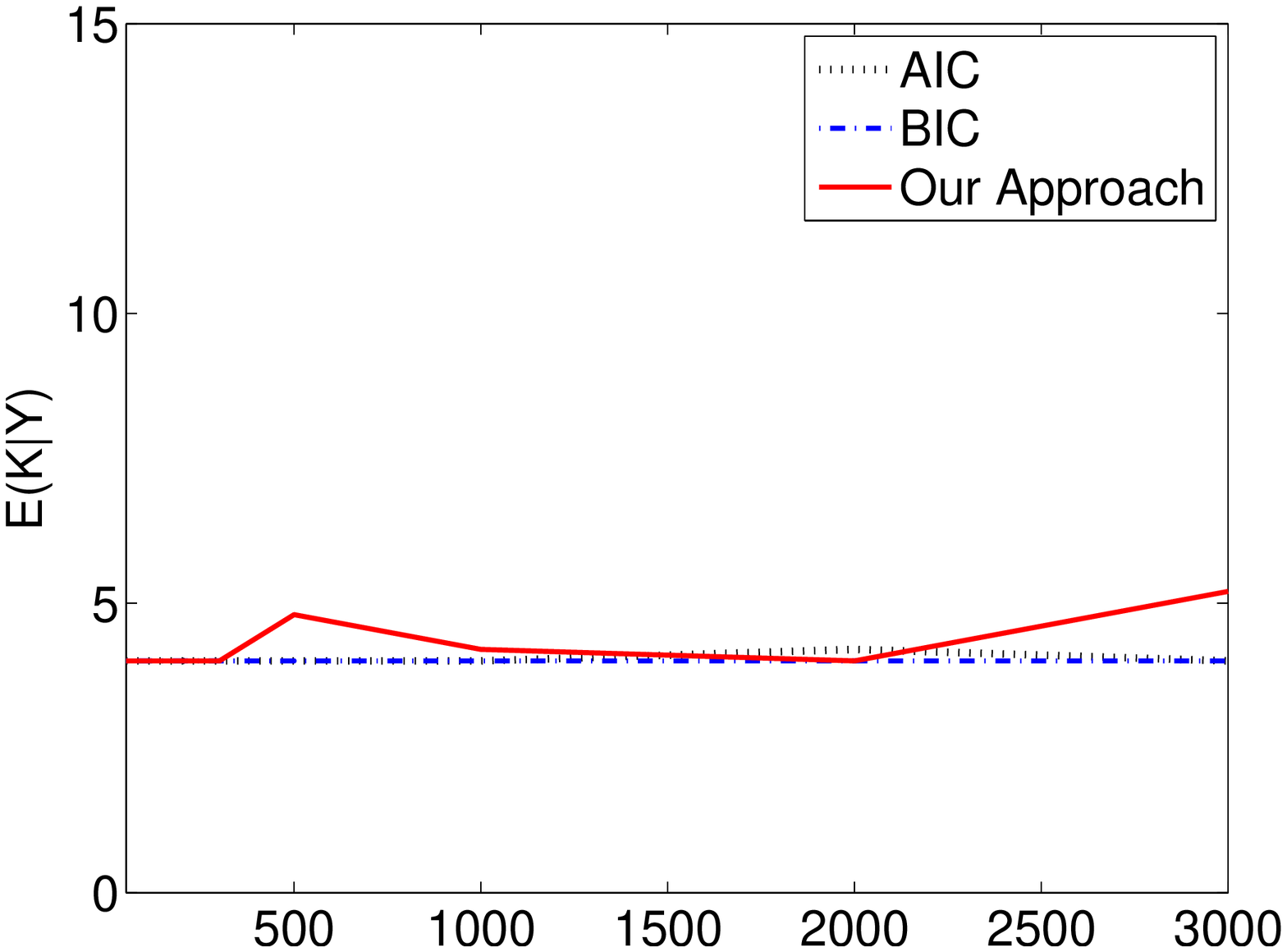}& 
\includegraphics[scale=0.145]{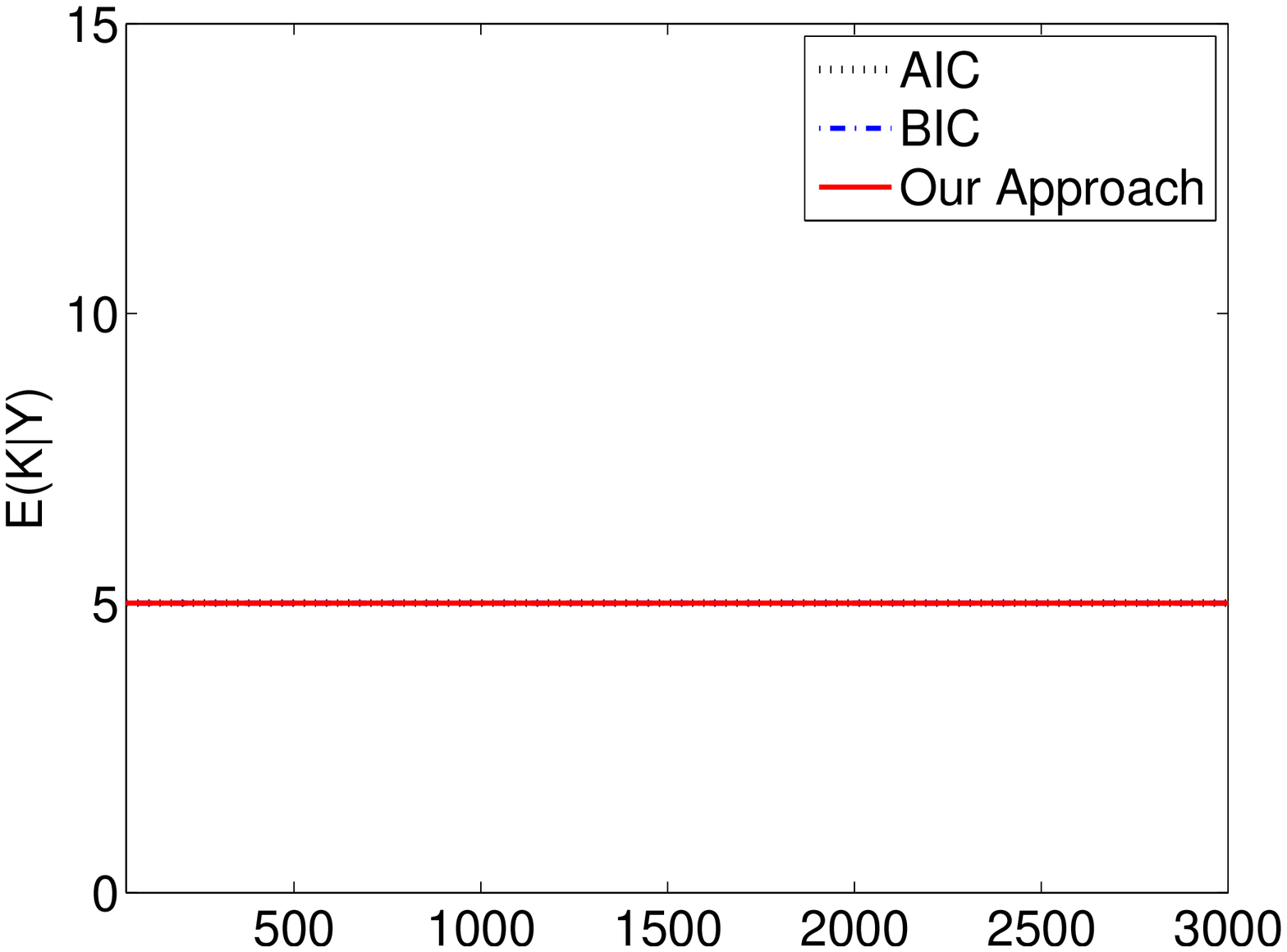}\cr
(b) MSE &
\includegraphics[scale=0.145]{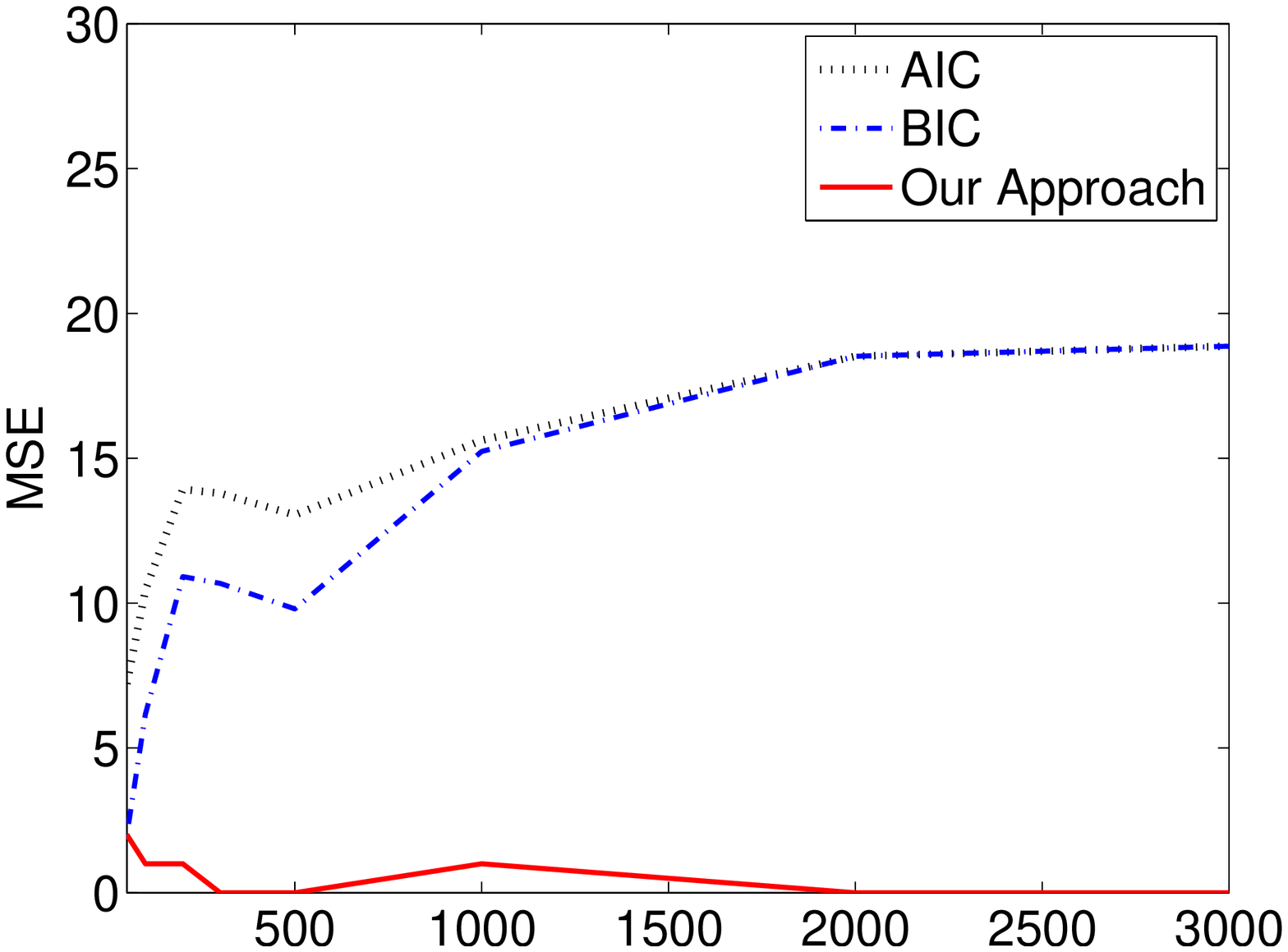}& 
\includegraphics[scale=0.145]{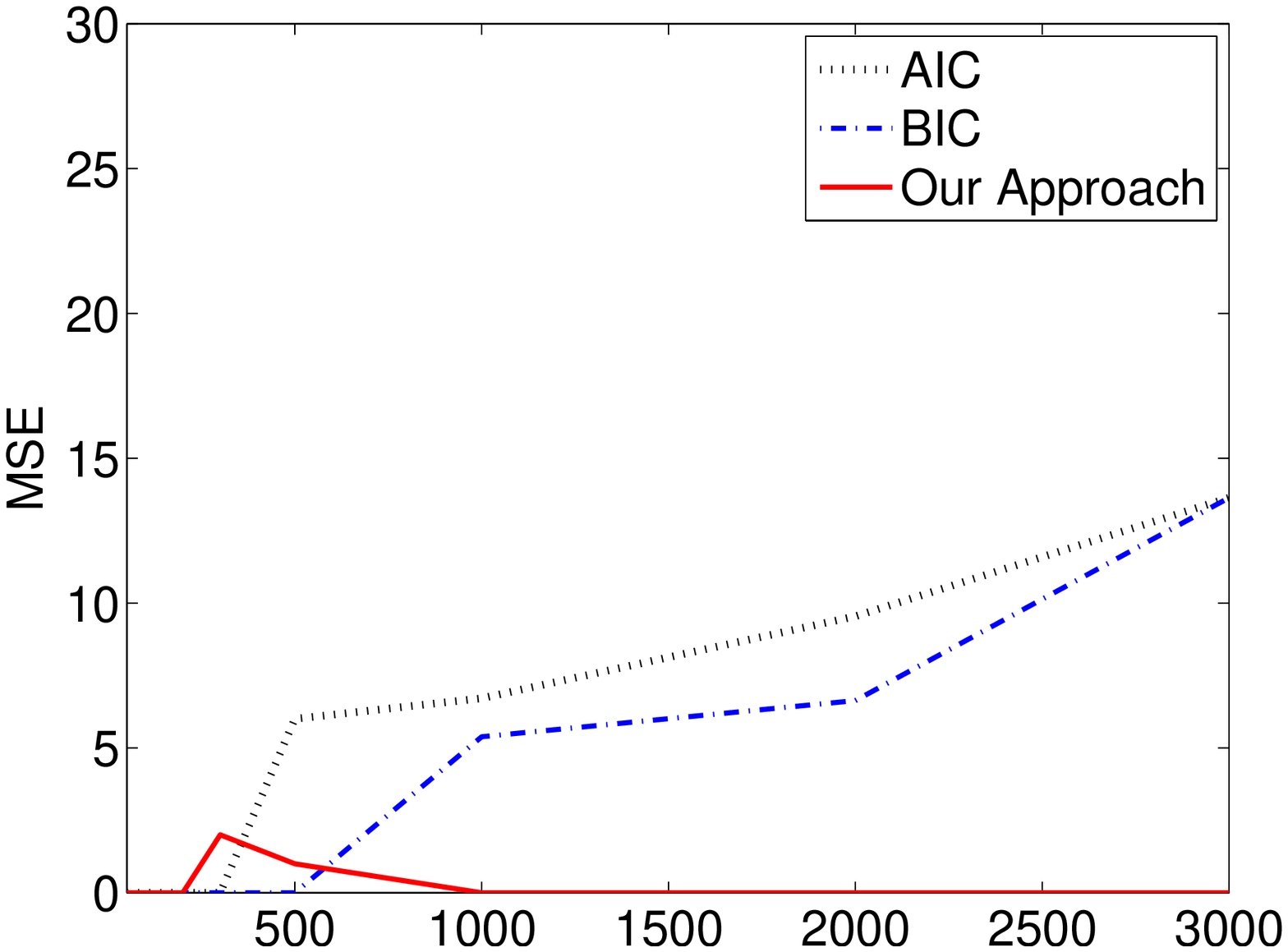}& 
\includegraphics[scale=0.145]{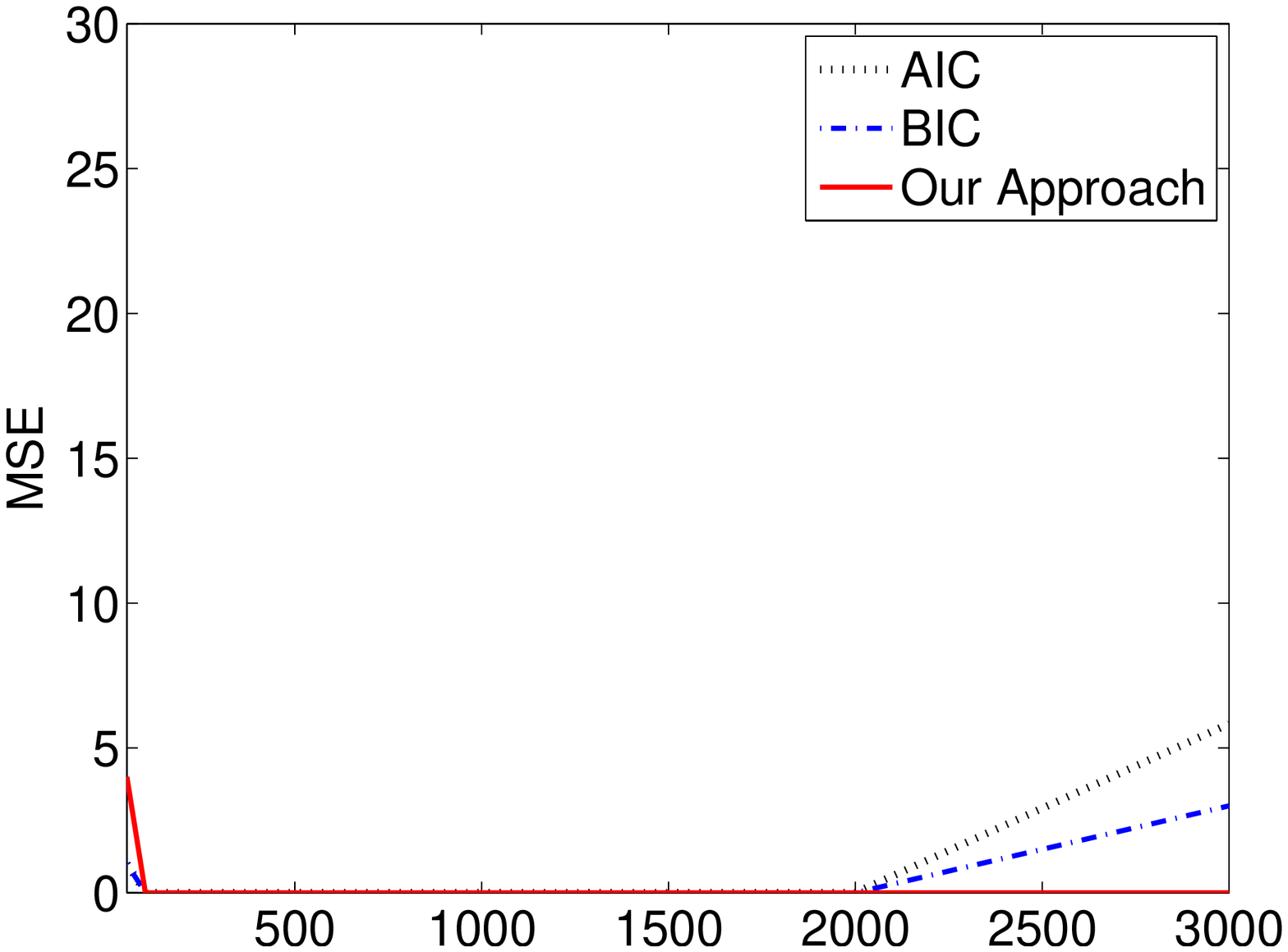}& 
\includegraphics[scale=0.145]{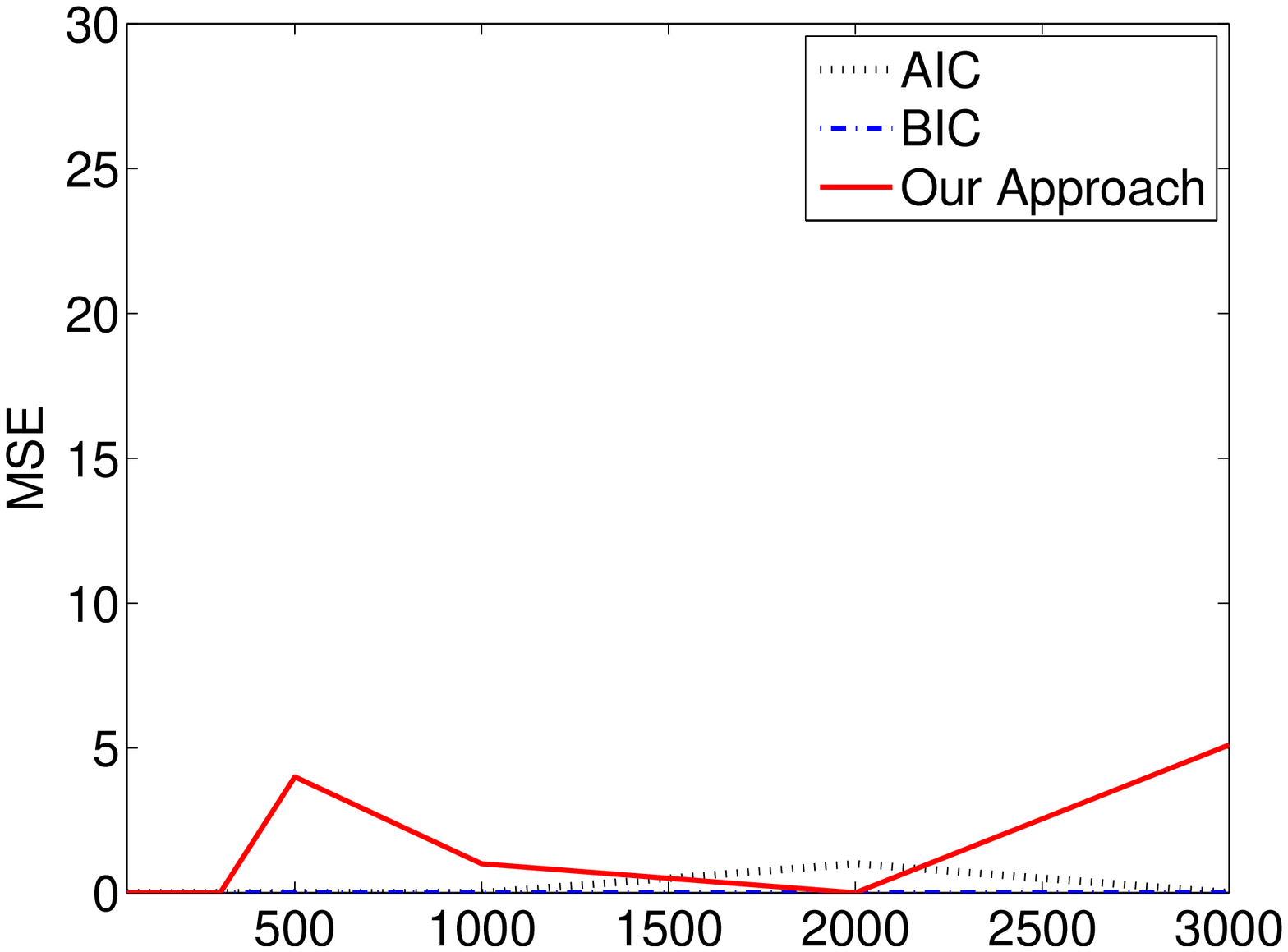}& 
\includegraphics[scale=0.145]{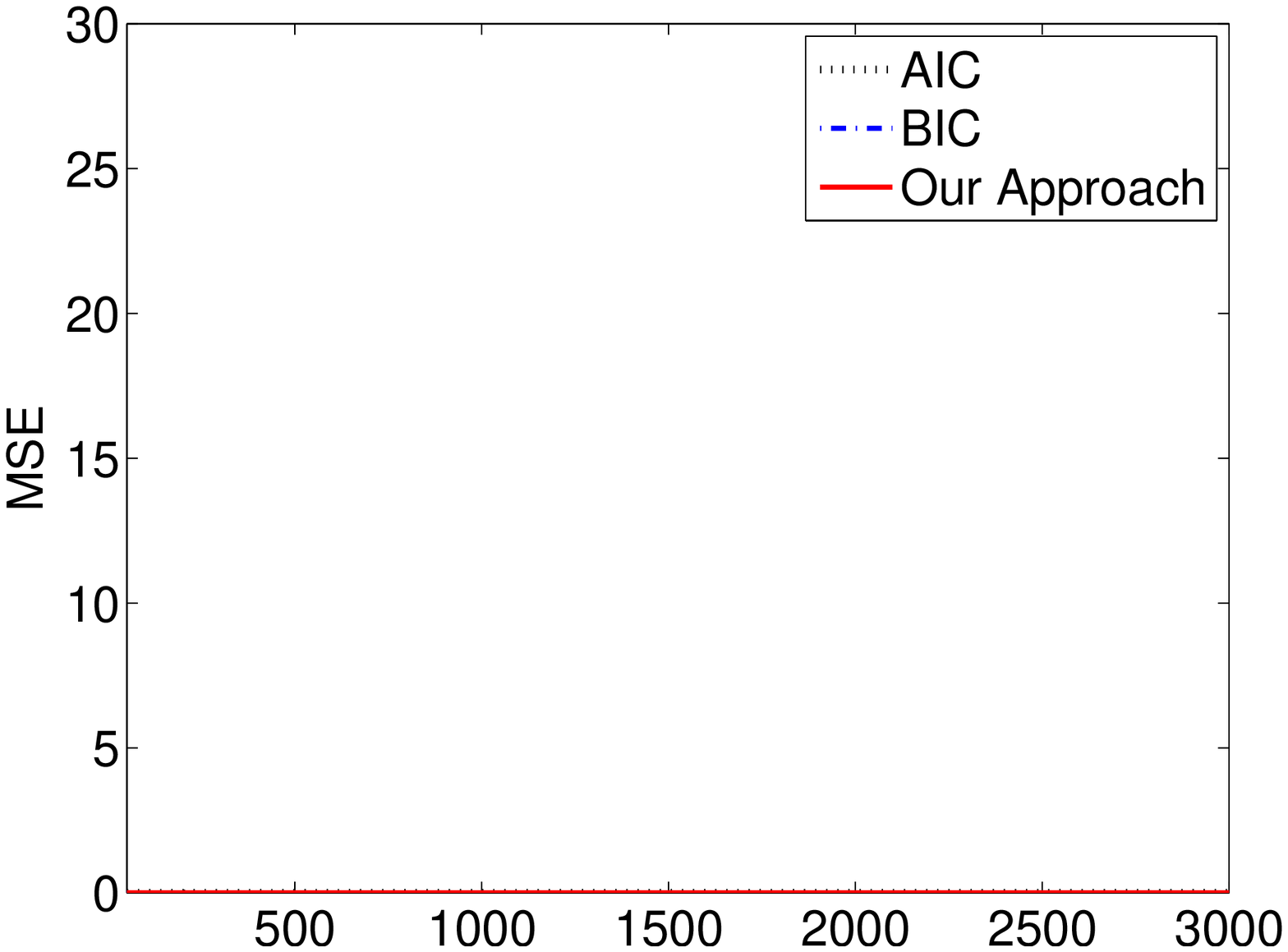}\cr
\end{tabular}
\caption{Mean of $K^{*}$ in five different runs and MSE with varying $\hat{K}$}
\label{fig: Mean of E(K|Y) and MSE with varying K}
\end{figure*}

We first investigated the performance of the proposed algorithm on two dimensional synthetic datasets for GMM clustering. Given $K$, data were generated by the hierarchical model:
\begin{eqnarray}
\mu_{j} &=& \left[ 20\cos\left( \frac{2\pi}{K}j \right), 20\sin\left(\frac{2\pi}{K}j \right) \right] \cr
\Sigma_{j} &\sim& \mathcal{W}({\bf I}_{2\times 2}, 5)  \textrm{ for } j\in\{1, 2, \cdots, K \} \cr
\pi_{1:K} &\sim & DP(1/K, 1/K, \cdots, 1/K)\cr
{\bf y}_{i} &\sim& \sum_{s=1}^{K} \pi_{s} \mathcal{N}(\cdot; \mu_{s}, \Sigma_{s}) \textrm{ for } i\in\{1, 2, \cdots, N \}
\end{eqnarray}
where $\mathcal{W}$ and $DP$ represent the Wishart distribution and Dirichlet Process respectively. In the experiments, we tested the performance by varying the number of clusters $\hat{K}\in\{1, 2, 3, 4, 5\}$ and the number of data $N\in\{50, 100, 200, 300, 500, 1000, 2000, 3000\}$. Figures \ref{fig: Clustered synthetic dataset with varying K} and \ref{fig: Mean of E(K|Y) and MSE with varying K} demonstrate a comparison of the performance of our approach with that of other model selection algorithms, AIC and BIC.

Figure \ref{fig: Clustered synthetic dataset with varying K} explains the outputs of three model selection approaches with different $\hat{K}$ and $N$. Interestingly, our proposed approach is effective even when $\hat{K}=1$, where both AIC and BIC fail. In addition, whereas AIC and BIC can find $K^{*}$ only when $N$ is large, our proposed approach builds a distinguishable and clear posterior distribution in all cases from $50$ to $3000$, and this enables us to detect the apparent $K^{*}$ close to $\hat{K}$ easily.
The next question that arises concerns the stability of our approach in noisy environments. Therefore, we ran five parallel and random simulations with different seeds. The mean and MSE (mean square error) of $K^{*}$ for five different runs are displayed in Figure \ref{fig: Mean of E(K|Y) and MSE with varying K}. We find that our proposed algorithm is stable even when $\hat{K}=1$ and $\hat{K}=2$, where AIC and BIC are not effective. Furthermore, AIC and BIC sharply increase the MSE (Mean Square Error) as $N$ increases, when $\hat{K}=1,2,3$. However, our approach has a small (close to zero) and stable MSE, although $N$ increases.

We also evaluated the performance of our algorithm with the three well-known data sets used in \cite{Richardson97:GMM} for real experimental data:  Enzymatic activity in the blood of 245 unrelated individuals \cite{Bechtel93:Mixture}, acidity in a sample of 155 lakes in the Northeastern United States \cite{Crawford94:Mixture}, and galaxy data with the velocities of 82 distant galaxies.
\begin{figure*}[ht!]
\centering
\begin{tabular}{ccc}
\includegraphics[width=2in, height=1in]{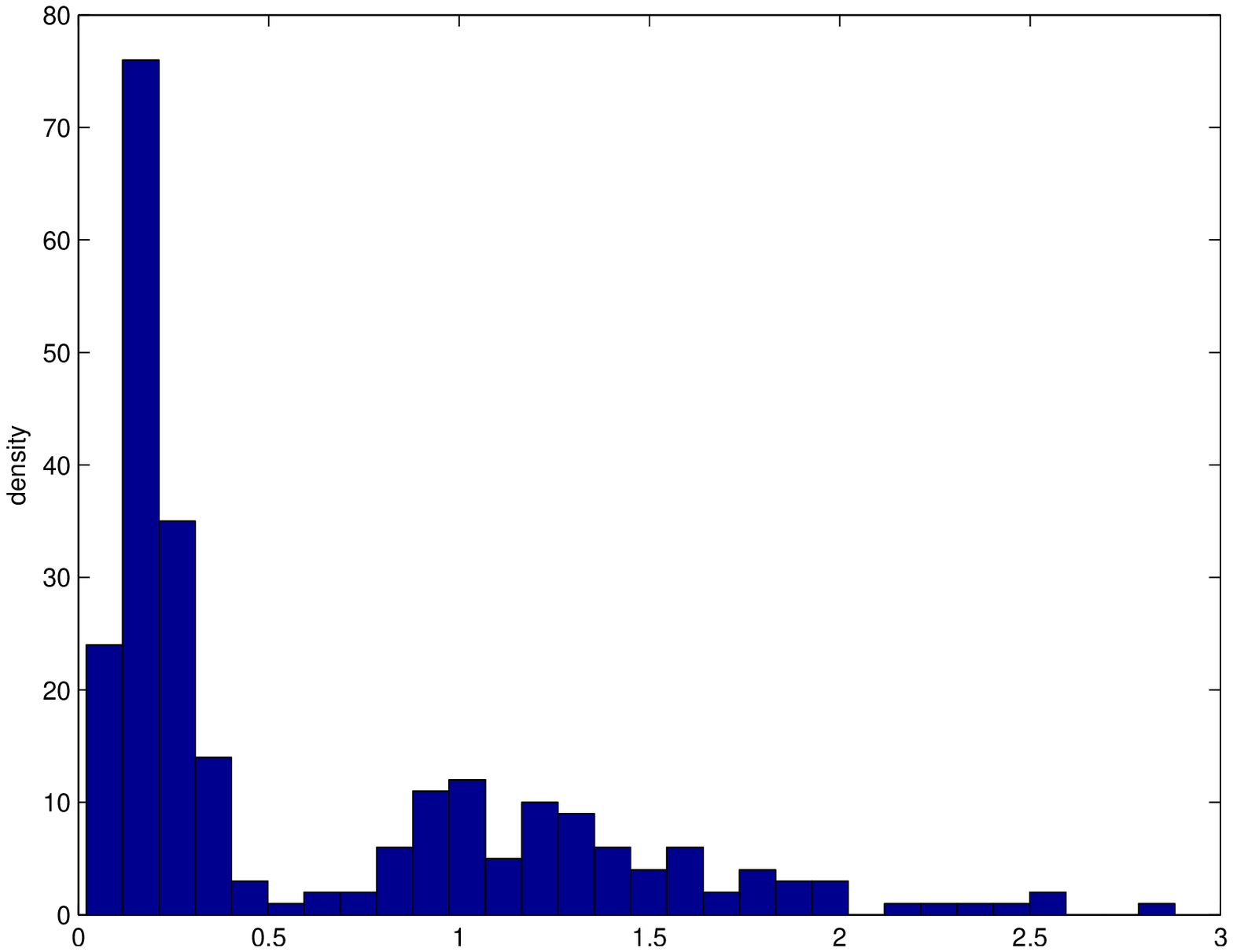}&
\includegraphics[width=2in, height=1in]{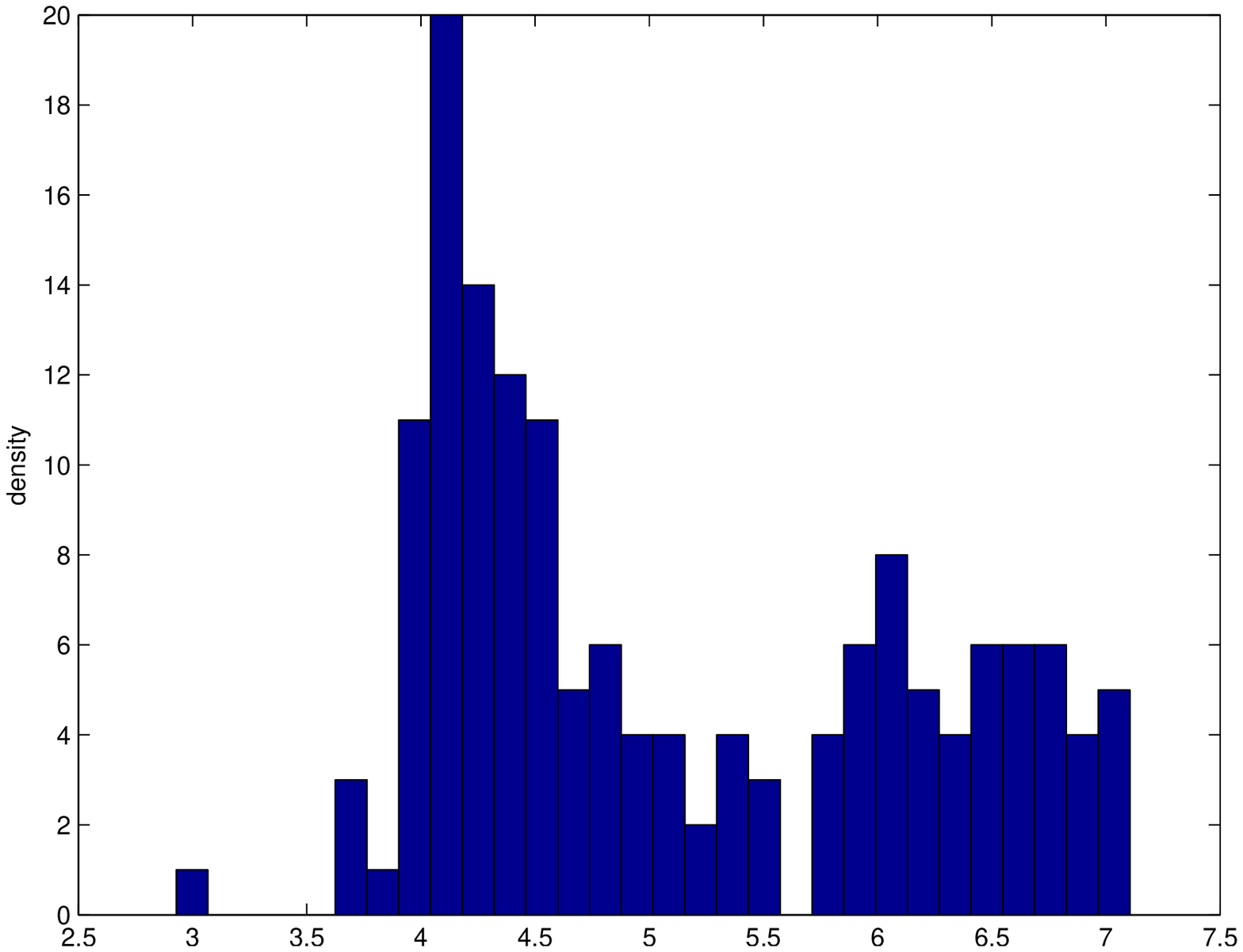}&
\includegraphics[width=2in, height=1in]{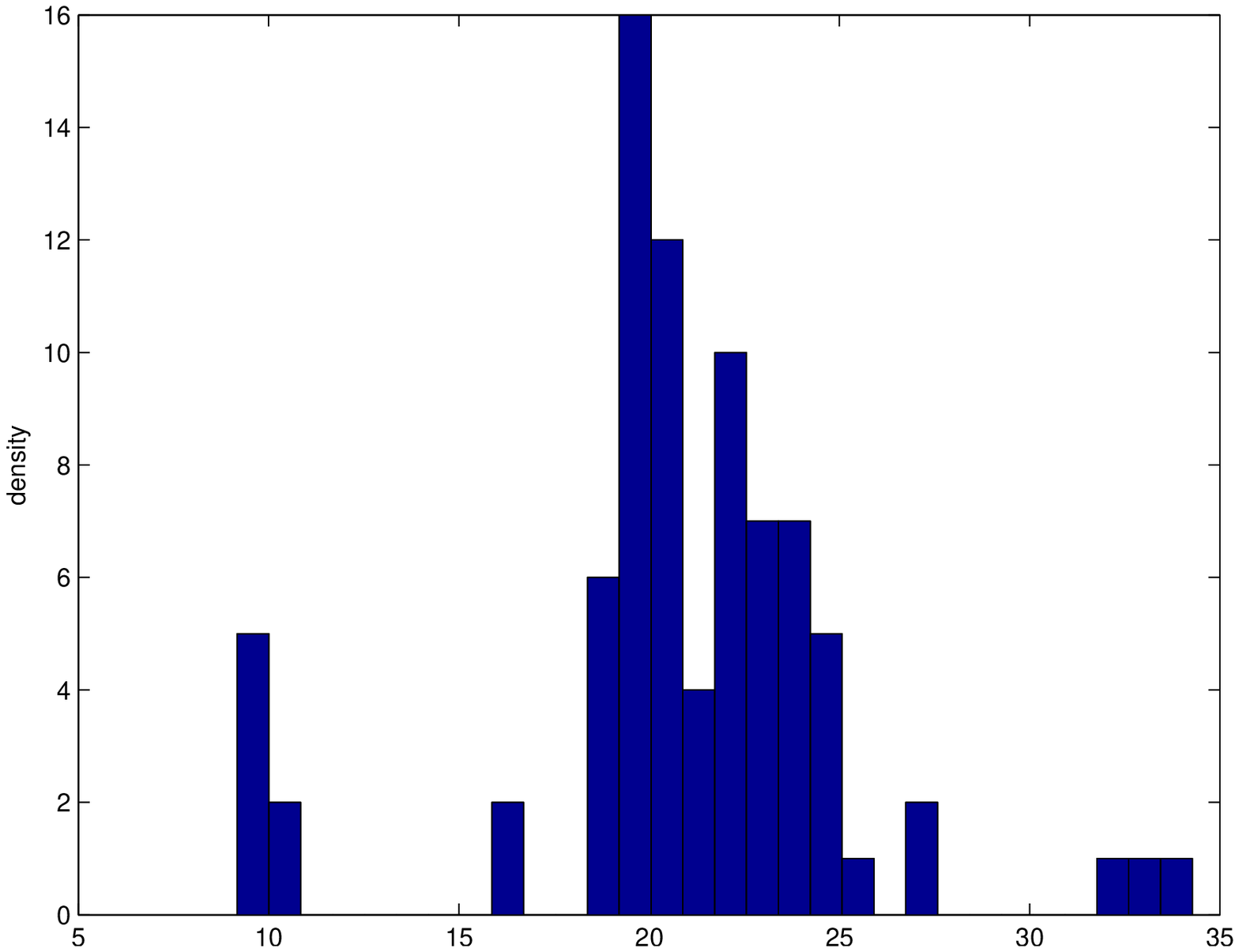}\cr
\includegraphics[width=2in, height=2in]{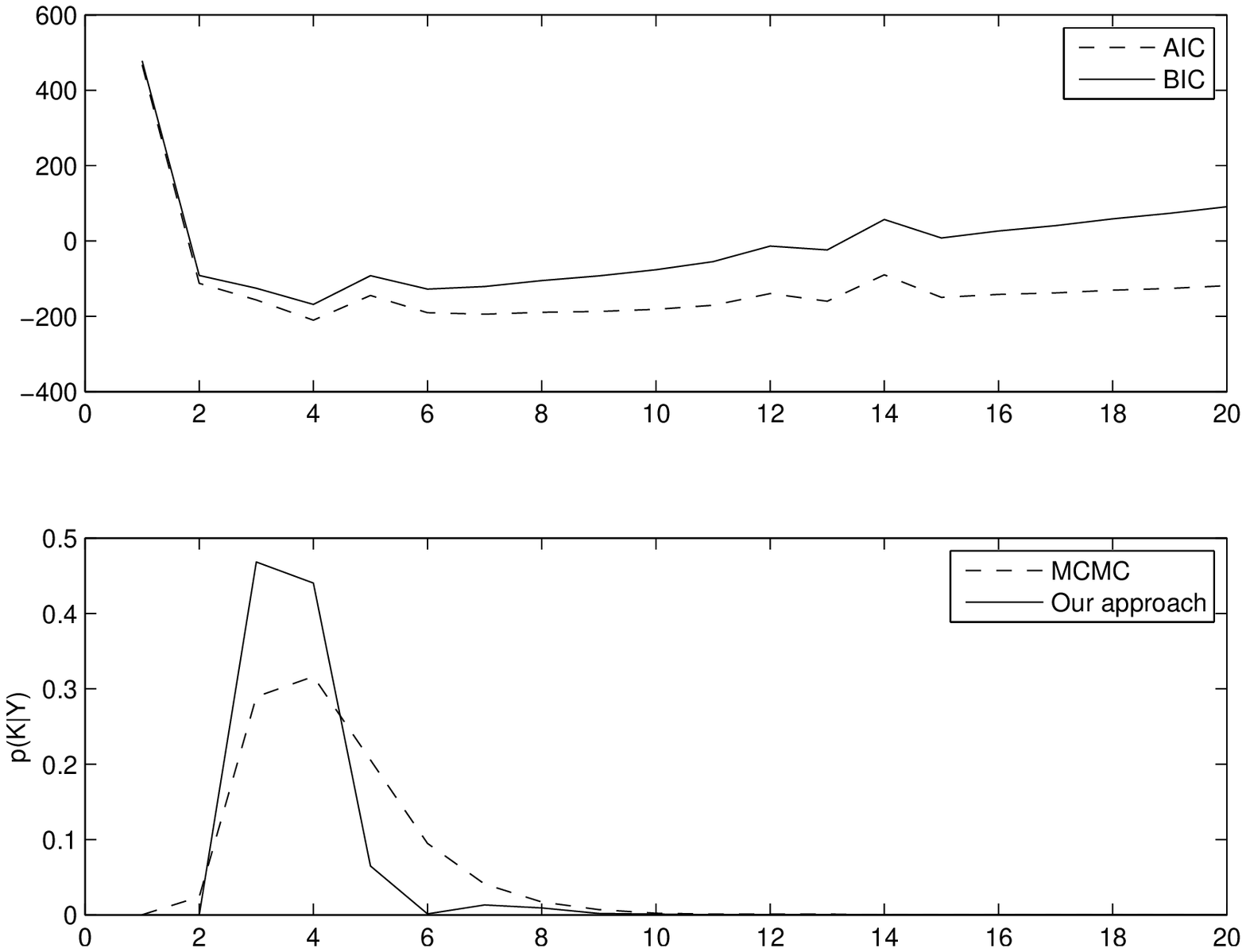}&
\includegraphics[width=2in, height=2in]{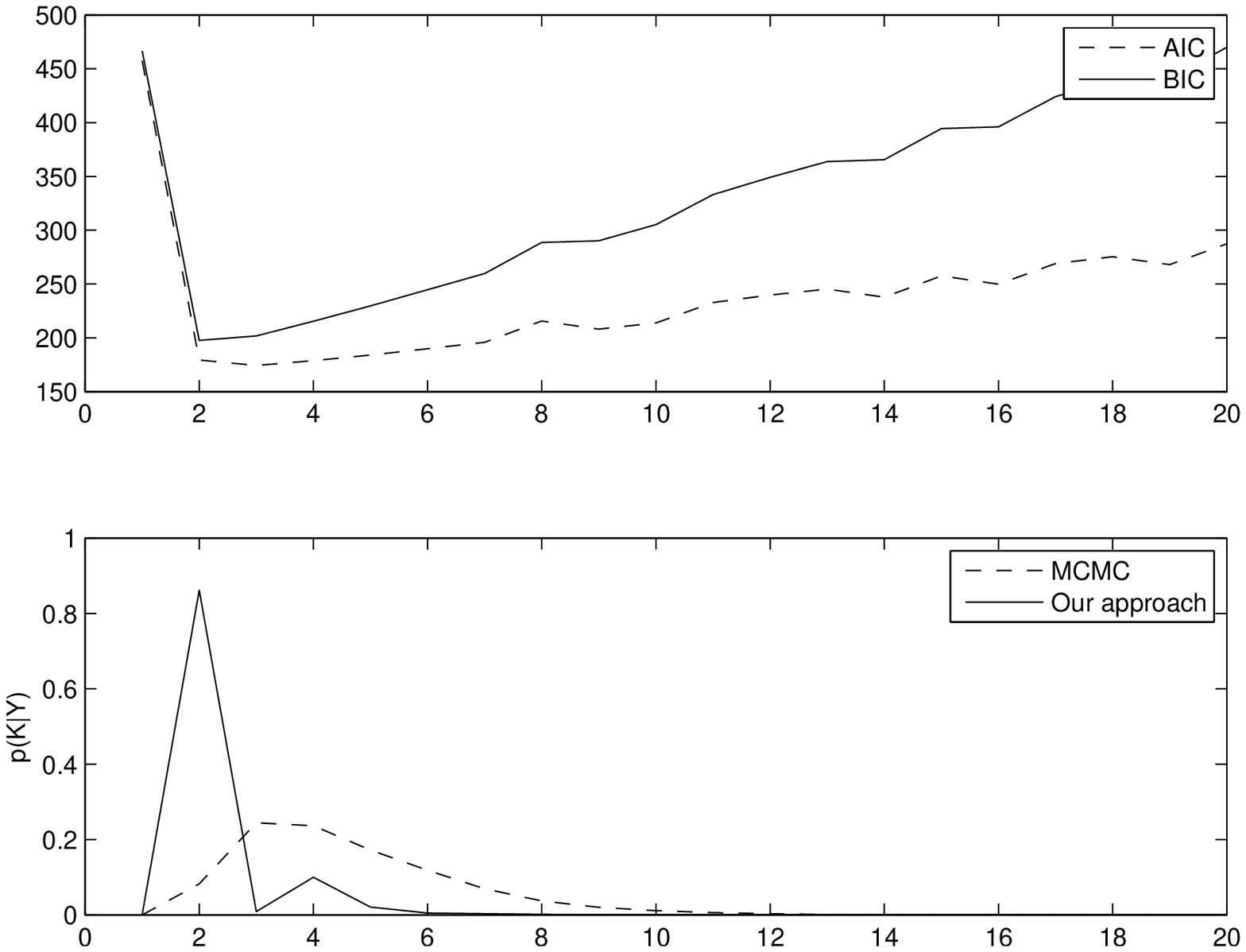}&
\includegraphics[width=2in, height=2in]{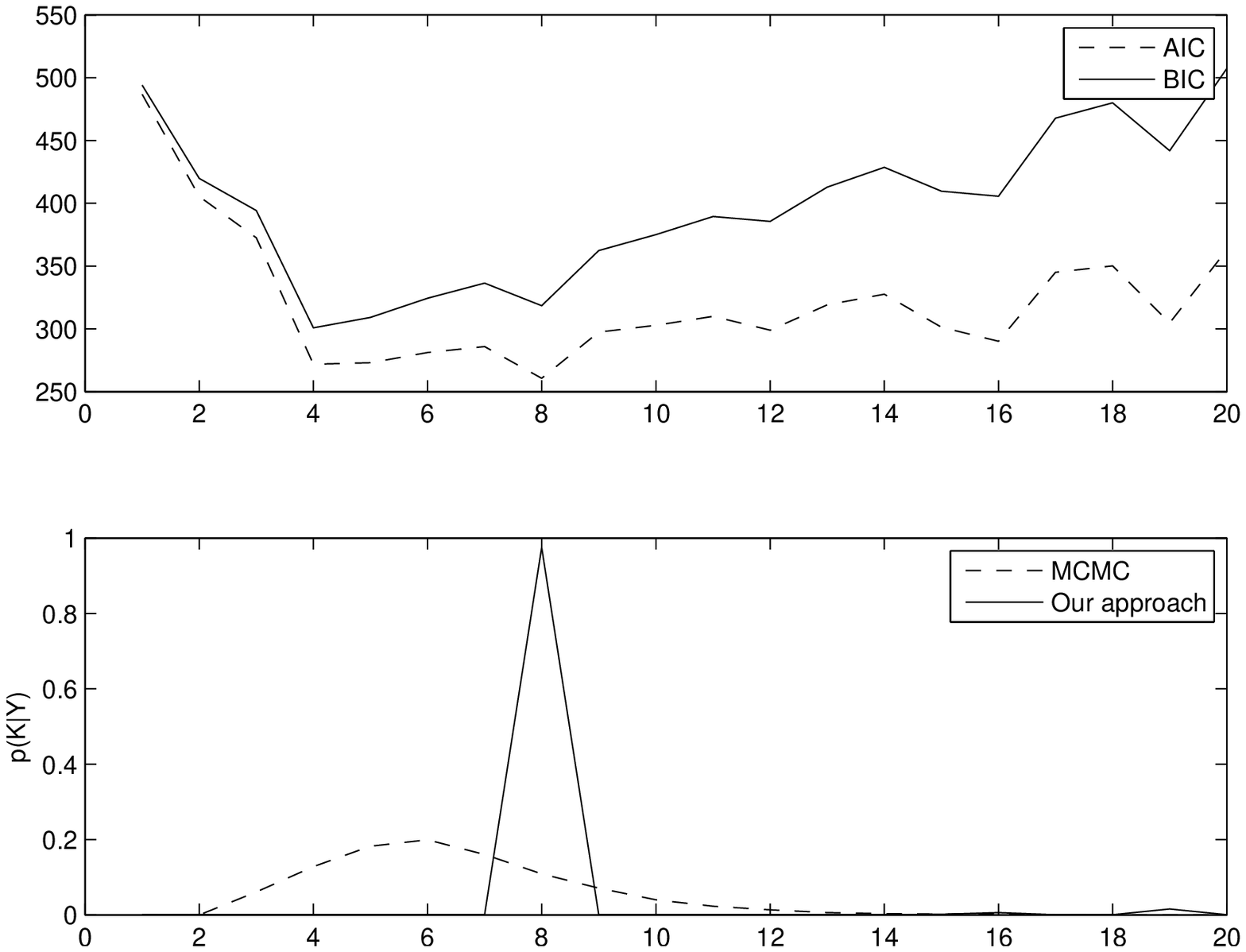}\cr
(a) Enzyme&
(b) Acidity&
(c) Galaxy
\end{tabular}
\caption{Histogram and its clustering results of one dimensional real datasets: (a) Enzyme (b) Acidity and (c) Galaxy}
\label{fig: Results of One dimensional real datasets}
\end{figure*}
Figure \ref{fig: Results of One dimensional real datasets} shows the performance comparison between AIC, BIC, MCMC, and our approach. The top sub-graphs demonstrate the histograms of the datasets with different numbers of mixture components. AIC and BIC with varying $K$ are plotted in the center row of sub-graphs. The bottom sub-figures display plots of the reconstructed distributions of $p(K|Y)$ by MCMC used in \cite{Richardson97:GMM} and our approach \footnote{The mathematical models used in MCMC and our approach are slightly different.}.

\section{Conclusion}
\label{section: Conclusion}

For Gaussian mixture clustering, we proposed a novel model selection algorithm, which is based on functional approximation in a Bayesian framework. This algorithm has a few advantages as compared to other conventional model selection techniques. First, the proposed approach can quickly provide a proper distribution of the model order which is not provided by other approaches, only a few time-consuming techniques such as Monte Carlo simulation can provide it. In addition, since the proposed algorithm is based on the Bayesian scheme, we do not need to run a cross validation, as is usually done in performance evaluation.

\bibliography{refs}
\bibliographystyle{plain}





\end{document}